\documentclass[acmsmall,screen,preprint, nonacm]{acmart}
\usepackage{amsfonts}
\usepackage{amsmath}
\usepackage{multirow}
\usepackage{graphicx}
\usepackage{subfig}
\usepackage{bm}

\usepackage{booktabs}
\usepackage{longtable}
\usepackage{arydshln} 
\usepackage{tikz}
\usepackage{forest}
\definecolor{orange}{rgb}{1,0.6,0}
\definecolor{midNightBlue}{rgb}{0.01,0.01, 0.55}

\usepackage[normalem]{ulem}
\usetikzlibrary{trees,positioning,shapes,shadows,arrows.meta}

\AtBeginDocument{%
  \providecommand\BibTeX{{%
    \normalfont B\kern-0.5em{\scshape i\kern-0.25em b}\kern-0.8em\TeX}}}

\copyrightyear{2025}
\acmYear{2025}

\begin{document}

\title{Compositional Zero-Shot Learning: A Survey}

\author{Ans Munir}
\email{msds20033@itu.edu.pk}
\affiliation{%
  \institution{Information Technology University}
  \city{Lahore}
  \state{Punjab}
  \country{Pakistan}
}

\author{Faisal Z. Qureshi}
\email{faisal.qureshi@ontariotechu.ca}
\affiliation{%
  \institution{University of Ontario Institute of Technology}
  \city{Oshawa}
  \state{Ontario}
  \country{Canada}
}

\author{Mohsen Ali}
\email{mohsen.ali@itu.edu.pk}
\affiliation{%
  \institution{Information Technology University}
  \city{Lahore}
  \state{Punjab}
  \country{Pakistan}
}

\author{Muhammad Haris Khan}
\email{muhammad.haris@mbzuai.ac.ae}
\affiliation{%
  \institution{Mohamed Bin Zayed University of Artificial Intelligence}
  \city{Abu Dhabi}
  \country{United Arab Emirates}
}


\begin{abstract}
\section*{Abstract}
Compositional Zero-Shot Learning (CZSL) is a critical task in computer vision that enables models to recognize unseen combinations of known attributes and objects during inference, addressing the combinatorial challenge of requiring training data for every possible composition. This is particularly challenging because the visual appearance of primitives is highly contextual; for example, ``small'' cats appear visually distinct from ``older'' ones, and ``wet'' cars differ significantly from ``wet'' cats. Effectively modeling this contextuality and the inherent compositionality is crucial for robust compositional zero-shot recognition. This paper presents, to our knowledge, the first comprehensive survey specifically focused on Compositional Zero-Shot Learning. We systematically review the state-of-the-art CZSL methods, introducing a taxonomy grounded in disentanglement, with four families of approaches: no explicit disentanglement, textual disentanglement, visual disentanglement, and cross-modal disentanglement. We provide a detailed comparative analysis of these methods, highlighting their core advantages and limitations in different problem settings, such as closed-world and open-world CZSL. Finally, we identify the most significant open challenges and outline promising future research directions. This survey aims to serve as a foundational resource to guide and inspire further advancements in this fascinating and important field. Papers studied in this survey with their official code are available on our github: \url{https://github.com/ans92/Compositional-Zero-Shot-Learning}
\end{abstract}

\keywords{}


\maketitle

\section{Introduction}
Zero-Shot Learning (ZSL) is a machine learning paradigm that allows models to classify objects from classes they were not seen during training. This is achieved by leveraging auxiliary information, such as textual descriptions or semantic embeddings, to bridge the gap between ``seen'' and ``unseen'' classes. The fundamental principle of ZSL is to transfer knowledge from a set of ``seen'' classes, for which labeled data is available, to make predictions about novel ``unseen'' classes. This capability is vital for overcoming the challenge of data scarcity, as it eliminates the need for exhaustive data collection and retraining for every new category. By doing so, ZSL makes artificial intelligence models more flexible and adaptable, allowing them to dynamically recognize new concepts.

\begin{figure}
    \centering
    \subfloat[]{\includegraphics[width=0.49\textwidth]{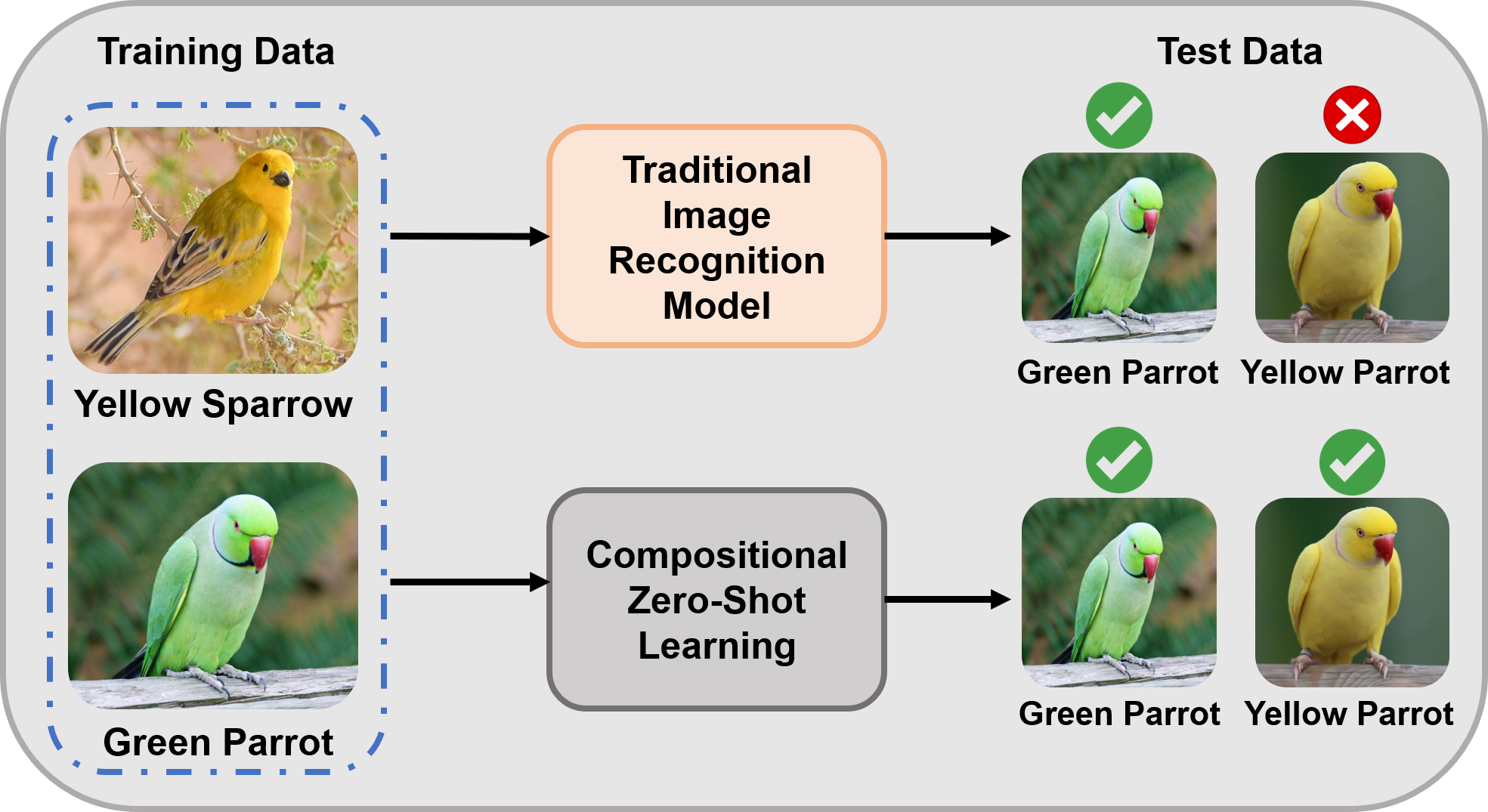}\label{fig:concept-diagram}}
  \hfill
  \subfloat[]{\includegraphics[width=0.48\textwidth, height=3.75cm]{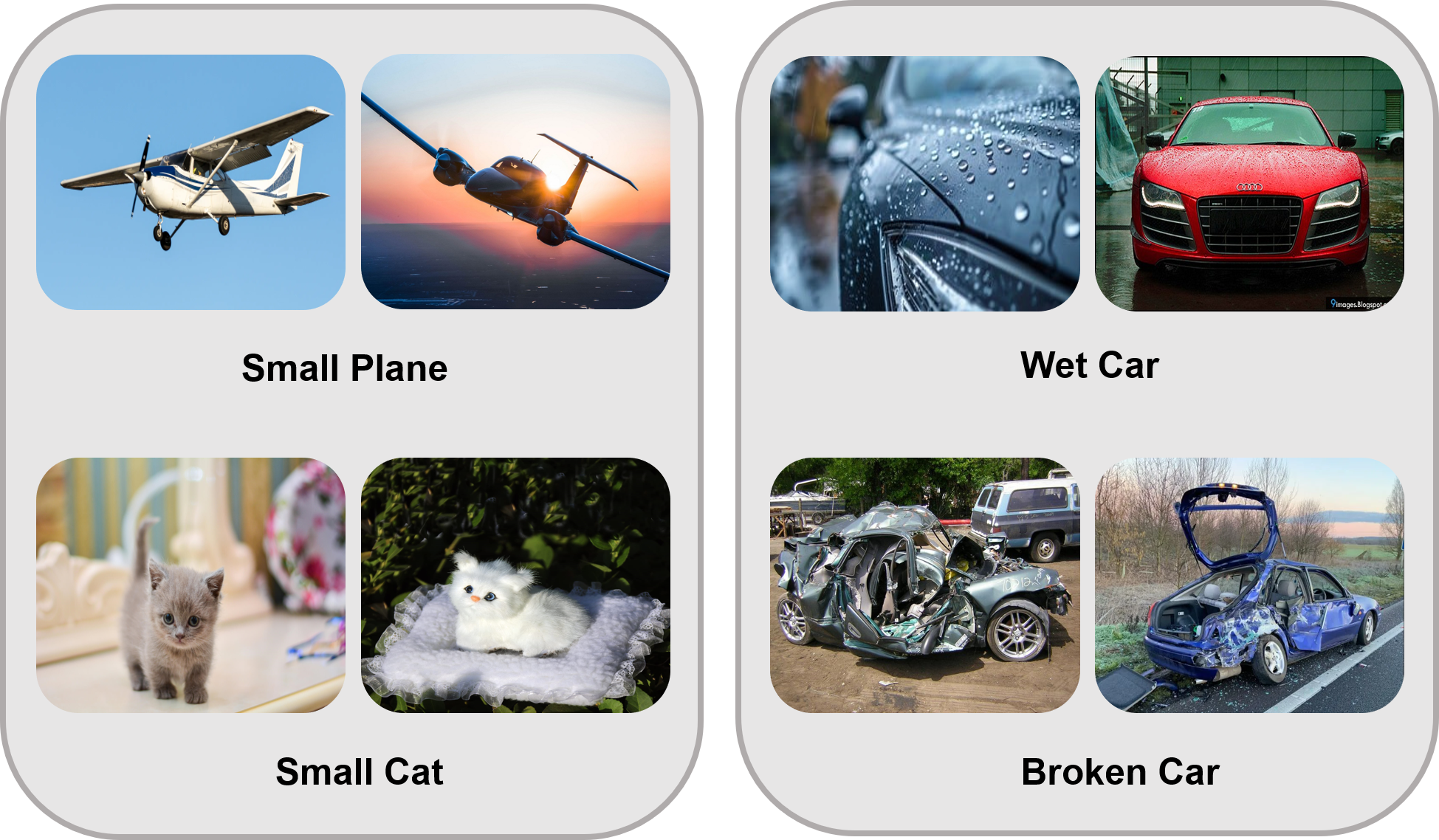}\label{fig:visual-diversity}}
  \caption{\textbf{(a)} Compositional Zero-Shot Learning concept diagram. Traditional Image Recognition Model is not able to understand and generalize concepts while Compositional Zero-Shot Learning models are able to generalize the attributes and objects to recognize unseen composition of seen concepts at test time. \textbf{(b)} Figure illustrates the importance of \textbf{contextuality} in CZSL. It shows how the attribute \emph{Small} appears differently when paired with different objects (\emph{Small Plane} vs. \emph{Small Cat}), and how the object \emph{Car} appears differently with different attributes (\emph{Wet Car} vs. \emph{Broken Car}).}
\end{figure}

One interesting zero-shot problem  is Compositional Zero-Shot Learning (CZSL) \cite{misra2017red, naeem2021learning, asp, huang2024troika, Retrieval}, focused on recognizing novel combinations of previously seen concepts. More specifically in CZSL, each class is typically defined as a composition of an object and its associated attribute, representing object's characteristics or state. In a zero-shot task in CZSL, given a set of combinations like `Green Parrot', `Yellow Sparrow', and `Red Car', the goal is to identify previously unseen combinations, such as `Yellow Parrot' (Fig. \ref{fig:concept-diagram}). CZSL is challenging because the number of compositions (attribute-object combinations) grows exponentially with the number of attributes and objects. Addressing this challenge requires more expressive and nuanced representations that can effectively disentangle and recombine the semantics of objects and attributes. The difficulty of learning expressive representations is exacerbated by the fact that the characteristics and visual appearance of an attribute often change depending on the object, and vice versa. For example, a ``wet car'' vs ``broken car'' or  ``small plane'' vs ``small cat'' (Fig. \ref{fig:visual-diversity}). 

\begin{figure}[!tbp]
  \centering
  \subfloat[]{\includegraphics[width=0.5\textwidth]{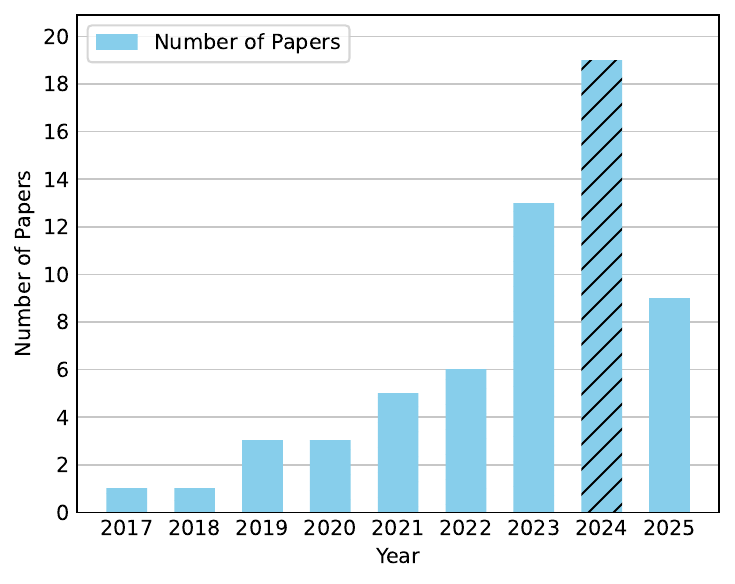}\label{fig:papers-by-year}}
  \hfill
  \subfloat[]{\includegraphics[width=0.5\textwidth]{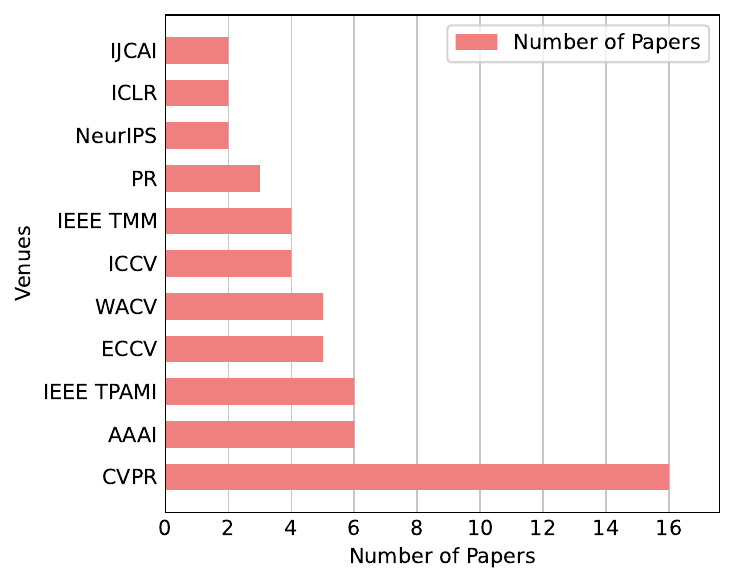}\label{fig:papers-by-venue}}
  \caption{Statistics of published work on CZSL problem in selected venues. (a) 
  Yearly breakdown of papers dealing with CZSL. As illustrated by the graph, CZSL papers saw a significant surge in 2023 and 2024. Specifically, 2023 marked a 116\% increase in publications over 2022, followed by a 46\% rise in 2024 compared to 2023. It should be noted that the 2025 publications comprise only those released during the first half of the year. (b) CZSL papers published from 2017 to 2025 in different venues.}
\end{figure}

Compositional Zero-Shot Learning has emerged as a field of significant promise, attracting significant research attention in recent years. This growing interest is unequivocally evidenced by the increasing volume of publications (Fig. \ref{fig:papers-by-year}) and the enhanced prominence of CZSL works in top-tier venues (Fig. \ref{fig:papers-by-venue}). These statistics were collected through a systematic search on Google Scholar using keywords such as `\textit{compositional zero-shot learning}', `\textit{state-object composition}', and `\textit{attribute-object recognition}'. To ensure quality and relevance, we restricted our analysis to (i) leading computer vision and machine learning conferences (e.g., CVPR, ICCV, ECCV, NeurIPS, ICLR, AAAI) and (ii) journals with an impact factor greater than 7 (e.g., IEEE TPAMI, Pattern Recognition (PR), IEEE TMM). This selection captures the most influential works in the field, and the resulting figures reflect both the publication volume and the distribution of CZSL research across these high-impact venues. For clarity, the abbreviations used in the Fig. \ref{fig:papers-by-venue} are expanded as follows: CVPR (Conference on Computer Vision and Pattern Recognition); ICCV (International Conference on Computer Vision); ECCV (European Conference on Computer Vision); NeurIPS (Neural Information Processing Systems); ICLR (International Conference on Learning Representations); AAAI (AAAI Conference on Artificial Intelligence); IJCAI (International Joint Conference on Artificial Intelligence); WACV (Winter Conference on Applications of Computer Vision); IEEE TPAMI (IEEE Transactions on Pattern Analysis and Machine Intelligence); PR (Pattern Recognition); IEEE TMM (IEEE Transactions on Multimedia).

While this area has rapidly evolved with diverse methods developed to tackle its inherent challenges, the burgeoning landscape of CZSL approaches currently lacks a foundational, comprehensive survey paper. 
This absence leaves a critical need for a systematic and fine-grained understanding of the underlying methodological philosophies and their evolution. Specifically, a structured categorization that articulates how various methods balance the fundamental requirements of learning distinct primitive representations versus modeling their contextual interactions has yet to be explored.

This survey addresses this critical gap by providing the \textbf{first in-depth} analytical framework and comprehensive review for categorizing CZSL methods. Our key contributions are as follows:
\begin{itemize}
    \item \textbf{A Comprehensive Taxonomy:} We introduce the first comprehensive taxonomy of Compositional Zero-Shot Learning (CZSL) methods, structured around the principle of disentanglement. At the first level, we categorize methods according to whether they explicitly disentangle primitives in the textual space, the visual space, both modalities, or not at all. At the second level, we further refine this organization by clustering methods according to their different strategies for modeling attributes and handling compositional challenges. This hierarchical structure provides a precise and non-overlapping organization of the literature, offering new insights into methodological design choices and the evolution of progress in CZSL.
    \item \textbf{Trend Identification and Performance Analysis:} We identify key performance trends and highlight the evolving dominance of different methodological paradigms, particularly the emergence and success of cross-modal (hybrid) approaches.
    \item \textbf{Identification of Open Challenges and Future Directions:} We consolidate and articulate the persistent challenges faced by current CZSL models and propose promising avenues for future research, guiding researchers towards impactful contributions in the field.
\end{itemize}

The remainder of this paper is structured as follows: Section \ref{sec:probl-seting-and-formulation} formulates the CZSL problem and discusses its different inference settings, including closed-world and open-world paradigms. Section \ref{sec:taxonomy} presents a systematic taxonomy of state-of-the-art CZSL methods, categorizing them according to their reliance on disentanglement (none, textual, visual, or both) and further refining them by their strategies for modeling attributes and addressing compositional challenges. Section \ref{sec:datasets} then reviews the benchmark datasets for CZSL, detailing their coverage of attributes and objects and the diversity of compositions, while Section \ref{sec:evaluation} summarizes the evaluation protocols and metrics that define closed-world and open-world settings. Building upon these foundations, Section \ref{sec:performance} provides a comparative analysis of existing methods, highlighting their key strengths, limitations, and empirical performance. Finally, Section \ref{open-challenges-future-direction} discusses the most significant open challenges and outlines promising directions for future research in CZSL.

\section{Problem Settings and Formulation in CZSL}
\label{sec:probl-seting-and-formulation}

\subsection{Problem Settings}

\begin{figure}
    \centering
    \includegraphics[width=0.9\linewidth]{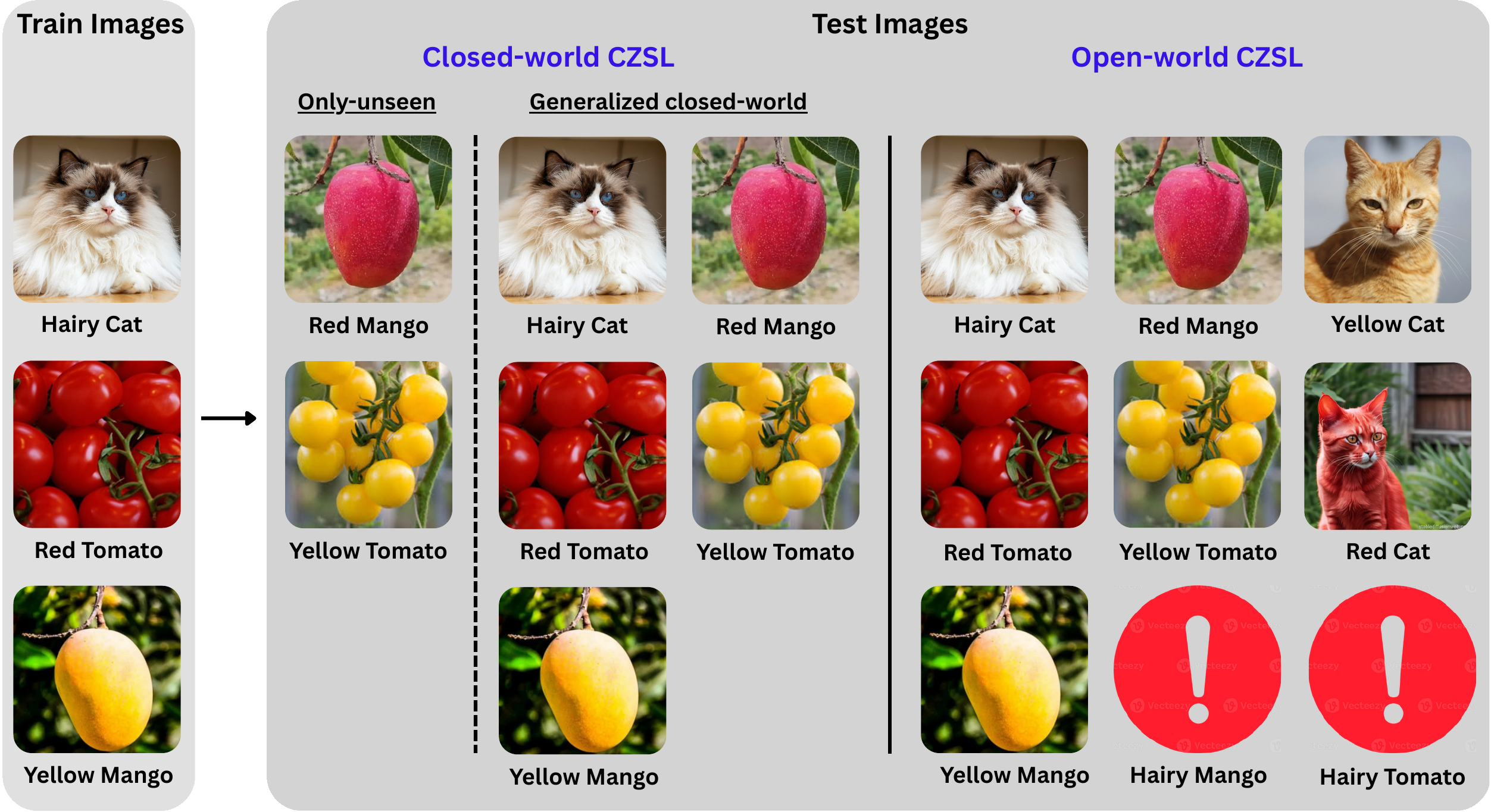}
    \caption{Difference between \textbf{Closed-world CZSL} and \textbf{Open-world CZSL} setting in Compositional Zero-Shot Learning. Closed-world setting is further divided into \textit{Only-unseen} and \textit{Generalized closed-world}. Compositions marked with exclamation marks indicate unfeasible compositions.} 
    \label{fig:close-world-open-world}
\end{figure}

Compositional Zero-Shot Learning (CZSL) is typically evaluated in two primary settings: (1) \textbf{Closed-world} and (2) \textbf{Open-world}. Traditionally, research has often focused on the \textit{closed-world setting}. 
In this scenario, the set of possible compositions, that the model might encounter during testing, are predefined and known. 
Within the closed-world setting, there are two common evaluation cases: (a) \textit{Only-unseen} Compositions where the model is presented only with compositions it did not see during training. (b) \textit{Generalized closed-world} where the model encounter instances from both the compositions it was trained on (seen compositions) and entirely new, unseen compositions. Fig. \ref{fig:close-world-open-world} illustrates the \textit{only-unseen} and \textit{generalized closed-world} situation. Suppose the training set includes compositions like ``Hairy Cat,'' ``Red Tomato,'' and ``Yellow Mango,'' along with corresponding images. During the testing phase, two unseen compositions, ``Red Mango'' and ``Yellow Tomato,'' along with their images may occur. In the \textit{generalized closed-world} case, objective of the model is to correctly recognize ``Yellow Tomato'' from a test set containing all five compositions (the original three seen and the two unseen). In contrast, in the \textit{only-unseen} case, the model would only be evaluated on instances of the two unseen compositions, ``Red Mango'' and ``Yellow Tomato,'' and its task is to predict ``Yellow Tomato'' from these two novel labels.  

An alternative evaluation scenario for CZSL is the open-world setting, first introduced by Mancini et al. \cite{mancini2021open}. Unlike the closed-world setting where the set of test compositions is known beforehand, the open-world scenario includes all possible combinations of attributes and objects in the test set. This comprehensive scope means the test set naturally contains compositions that might not exist or be physically possible in the real world. Fig. \ref{fig:close-world-open-world} provides an illustration within the open-world, where compositions like ``Hairy Tomato'' and ``Hairy Mango'' serve as examples of such unfeasible combinations. This expanded test space introduces two major challenges: firstly, the sheer size of the test set escalates dramatically as the number of attributes and objects grows, and secondly, the model must be robust enough to not only recognize valid compositions but also to avoid incorrectly predicting these unfeasible ones.

\subsection{Problem Formulation}


Let $\bm{A}$ be the set of $n$ attributes, denoted as $\bm{A} = \{ a_1, a_2, ..., a_n \}$ and $\bm{O}$ be the set of $m$ objects, denoted as $\bm{O} = \{ o_1, o_2, ..., o_m \}$ The set of all possible compositions is defined as the Cartesian product of the attribute and object sets, $\bm{C} = \bm{A} \times \bm{O}$. We consider the task of visual classification where each image $\bm{x}$ is associated with a compositional label $c \in \bm{C}$. Specifically, each compositional label $c$ is a pair $(a,o)$, where $a \in \bm{A}$ is an attribute primitive and $o \in \bm{O}$ is an object primitive.


The training data consists of a set of compositions $\bm{C}_{train} \subset \bm{C}$ and their corresponding images $\bm{X}_{train}$. The composition set used for evaluation at test time, denoted as $\bm{C}_{test}$, is defined in three principal ways within the CZSL literature. In the \textbf{only-unseen} setting, the test set consists only of novel compositions ($\bm{C}_{unseen}$, with $\bm{C}_{train} \cap \bm{C}_{unseen} = \emptyset $), meaning $\bm{C}_{test} \subset \bm{C}$ and $\bm{C}_{train} \cap \bm{C}_{test} = \emptyset $. In the \textbf{generalized closed-world} setting, the set of compositions the model must classify over at test time includes both the training compositions and the novel unseen compositions. This test set is often represented as $\bm{C}_{test} = \bm{C}_{train} \cup \bm{C}_{unseen}$. The third setting is the \textbf{open-world}, where the potential test set for classification encompasses all possible compositions, meaning $\bm{C}_{test} = \bm{C} = \bm{A} \times \bm{O}$.


\subsection{Motivation for the Taxonomy of CZSL Methods}
\label{sec:definitions}
While the task of Compositional Zero-Shot Learning (CZSL) is unified by its goal of recognizing novel attribute-object combinations, the rapid expansion of approaches has produced a diverse and fragmented landscape. Without a principled framework, it is difficult to compare methods, identify their underlying assumptions, or chart promising directions for future research. To address this, we organize existing methods around the central principle of feature disentanglement, which is the degree to which models explicitly separate and represent the constituent primitives of a composition (attributes and objects) so that they can later be flexibly recombined to recognize unseen combination.

Disentanglement serves as a natural basis for organizing CZSL methods, since it directly addresses the two fundamental challenges of the task: discriminability and contextuality. \textbf{Discriminability} refers to the ability of a model to learn distinct and robust representations for individual primitives. Without disentanglement, models tend to conflate correlations between attributes and objects observed during training. For example, an excessive association of ``red'' with ``tomato'' can bias predictions for unseen pairs. By contrast, disentangled representations enable the model to separate the concept of ``red'' from the concept of ``tomato,'' making it possible to generalize correctly to novel compositions such as ``red car.'' \textbf{Contextuality}, on the other hand, concerns how the expression of a primitive depends on the object it modifies. The same attribute may manifest in very different ways depending on its context; for example, ``small'' conveys a different scale when applied to a ``plane'' than to a ``cat.'' Effective disentanglement allows models to retain this flexibility, ensuring that primitives can adapt to new contexts rather than remaining tied to correlations observed during training.


With these challenges in view, we categorize existing methods into four broad groups based on where and how disentanglement is applied:

\begin{itemize}
    \item \textbf{No Explicit Disentanglement:} Methods in this category avoid explicit disentanglement and instead model attribute-object compositions through holistic embeddings or direct fusion mechanisms. While conceptually simpler, these approaches provide essential reference points for quantifying the value added by disentanglement-based frameworks.
    \item \textbf{Textual Disentanglement:} It involves separating the semantic embeddings of primitives within the language space. Through the acquisition of independent concept representations, models are able to systematically compose these elements, enabling generalization to unseen attribute-object combinations.
    \item \textbf{Visual Disentanglement:} These methods isolate the visual features of attributes and objects within image representations. By disentangling these primitives into composable representations, the models can recombine them to identify novel attribute-object compositions at test time.
    \item \textbf{Cross-Modal (Hybrid) Disentanglement:} These approaches simultaneously disentangle primitives in visual and textual spaces. Through cross-modal alignment, they integrate complementary information from both modalities, thereby enhancing the ability to compose reliable representations for novel attribute-object combinations.
\end{itemize}

These four disentanglement strategies constitute the core foundation of our proposed taxonomy, which is elaborated in the following section.


\begin{figure}
    \centering
    
\tikzset{
    basic/.style  = {draw, text width=1cm, align=center, font=\sffamily, rectangle},
    root/.style   = {basic, rounded corners=2pt, thin, align=center, fill=brown!30!orange!40!white, text width=2.2cm, minimum height=2.0cm, inner sep=2.2pt},
    onode/.style = {basic, thin, rounded corners=2pt, align=center, fill=green!60,text width=4.4cm,},
    tnode/.style = {basic, thin, rounded corners=2pt, fill=pink!60, text width=18.3em, align=center},
    xnode/.style = {basic, thin, rounded corners=2pt, align=center, fill=blue!20,text width=2.4cm},
    wnode/.style = {basic, thin, align=left, fill=pink!10!blue!80!red!10, text width=6.5em},
    edge from parent/.style={draw=black, edge from parent fork right}

}
{\footnotesize
\begin{forest} for tree={
    grow=east,
    growth parent anchor=west,
    parent anchor=east,
    child anchor=west,
    edge path={\noexpand\path[\forestoption{edge},-, >={latex}] 
         (!u.parent anchor) -- +(10pt,0pt) |-  (.child anchor) 
         \forestoption{edge label};}
}
[{Disentanglement in CZSL}, root,  l sep=6mm,
    [Cross-Modal (Hybrid) Disentanglement, xnode,  l sep=5mm, fill=purple!9!white
        [Encoder-Augmented Modeling, xnode, text width=2.6cm, fill=purple!5!white
        [CAILA \cite{caila}; Troika \cite{huang2024troika} , onode, fill=gray!3!white]]
        [LLM-Guided Prompting, xnode, text width=2.6cm, fill=purple!5!white
        [PLID \cite{PLID} , onode, fill=gray!3!white]]
        ]
    [{Visual Feature Disentanglement}, xnode,  l sep=5mm, fill=blue!20!white
        [Axiomatic Modeling Approaches, xnode, text width=2.6cm, fill=blue!9!white
        [SymNet \cite{li2020symmetry}  , onode, fill=gray!3!white]]
        [Dual-Stream Visual Modeling, xnode, text width=2.6cm, fill=blue!9!white
        [ATIF \cite{min2024adaptive}  , onode, fill=gray!3!white]]
        [Synthetic Embedding Methods, xnode, text width=2.6cm, fill=blue!9!white
        [HiDC \cite{HiDC}; Generative \cite{nan2019recognizing}; OADis \cite{OADis}; SAD-SP \cite{SAD-SP}; LIDF \cite{LIDF}, onode, fill=gray!3!white]]
        [Conditional Attribute Modeling, xnode, text width=2.6cm, fill=blue!9!white
        [CANet \cite{CANet}; CDS-CZSL \cite{CDS-CZSL} , onode, fill=gray!3!white]]
        [Auxiliary Objective-Based Methods, xnode, text width=2.6cm, fill=blue!9!white
        [ADE \cite{ADE}; CCZSL \cite{zhang2024continualCZSL}; OV-CZSL \cite{OV-CZSL}; DRANet \cite{DRANet}; IVR \cite{IVR}; VisPrompt \cite{VisPrompt}; LVAR \cite{ma2024lvar} , onode, fill=gray!3!white]]
        [Prototype-Centric Methods, xnode, text width=2.6cm, fill=blue!9!white
        [CLUSPRO \cite{CLUSPRO}; HOPE \cite{HOPE}; Protoprop \cite{Protoprop} ; SCD \cite{SCD}; CLPSL \cite{CLPSL}   , onode, fill=gray!3!white]]
        [Augmentation-Enhanced Methods, xnode, text width=2.6cm, fill=blue!9!white
        [SCEN \cite{SCEN}; CoT \cite{CoT}; Retrieval \cite{Retrieval}; MSCC \cite{MSCC}  , onode, fill=gray!3!white]]
         ]
    [{Textual Feature Disentanglement}, xnode,  l sep=5mm, fill=green!20!white
        [Primitive-Aware Modeling, xnode, text width=2.6cm, fill=green!9!white
        [HPL \cite{HPL}; DFSP \cite{DFSP}, onode, fill=gray!3!white]]
        [Pairwise Composition Modeling, xnode, text width=2.6cm, fill=green!9!white
        [AoP \cite{nagarajan2018attributes}; BMP-Net \cite{BMP-Net}; ASP \cite{asp}; GIPCOL \cite{Gipcol} ; CatCom \cite{Catcom}, onode, fill=gray!3!white]]
         ]
    [No Explicit Disentanglement, xnode,  l sep=5mm, fill=gray!15!white 
        [Partial Composition Learning Models, xnode, text width=2.6cm, fill=gray!9!white 
        [KG-SP \cite{kg-sp}; ProCC \cite{huo2024procc} , onode, fill=gray!3!white]]
        [Multi-Domain CZSL Methods, xnode, text width=2.6cm, fill=gray!9!white 
        [Dynamic\_learning \cite{hu2024dynamic} , onode, fill=gray!3!white]]
        [Causality-Driven Methods, xnode, text width=2.6cm, fill=gray!9!white 
        [Causal \cite{atzmon2020causal}; DeCa \cite{DeCa} , onode, fill=gray!3!white]]
        [Constraint-Guided Learning, xnode, text width=2.6cm, fill=gray!9!white 
        [Adversarial \cite{wei2019adversarial}; Compcos \cite{naeem2021learning} , onode, fill=gray!3!white]]
        [Context-Aware Composition Modeling, xnode, text width=2.6cm, fill=gray!9!white 
        [ CGE \cite{naeem2021learning}; Co-CGE \cite{mancini2022learning}; CAPE \cite{CAPE}; ProLT \cite{ProLT} , onode, fill=gray!3!white]]
        [Direct Composition Modeling, xnode, text width=2.6cm, fill=gray!9!white 
        [RedWine \cite{misra2017red}; CSP \cite{csp2023}; TMN \cite{purushwalkam2019task} , onode, fill=gray!3!white]]
         ]]
\end{forest}
}
    \caption{A Comprehensive Taxonomy of Compositional Zero-Shot Learning Methods. Our taxonomy organizes CZSL methods along two dimensions. At the first level, methods are grouped by their disentanglement strategy, meaning how they factorize primitive representations across modalities. \textbf{No Explicit Disentanglement} retains unified representations of whole composition without modular separation; \textbf{Textual Feature Disentanglement} isolates attribute and object semantics within the language space; \textbf{Visual Feature Disentanglement} explicitly separates attribute-related and object-related components in the visual feature space; and \textbf{Cross-Modal (Hybrid) Disentanglement} factorizes both modalities jointly, aligning visual and textual primitives in a structured way. At the second level, methods are further categorized by their approach to solve CZSL challenge. Representative papers are listed under each category.}
    \label{fig:taxonomy-diagram}
\end{figure}
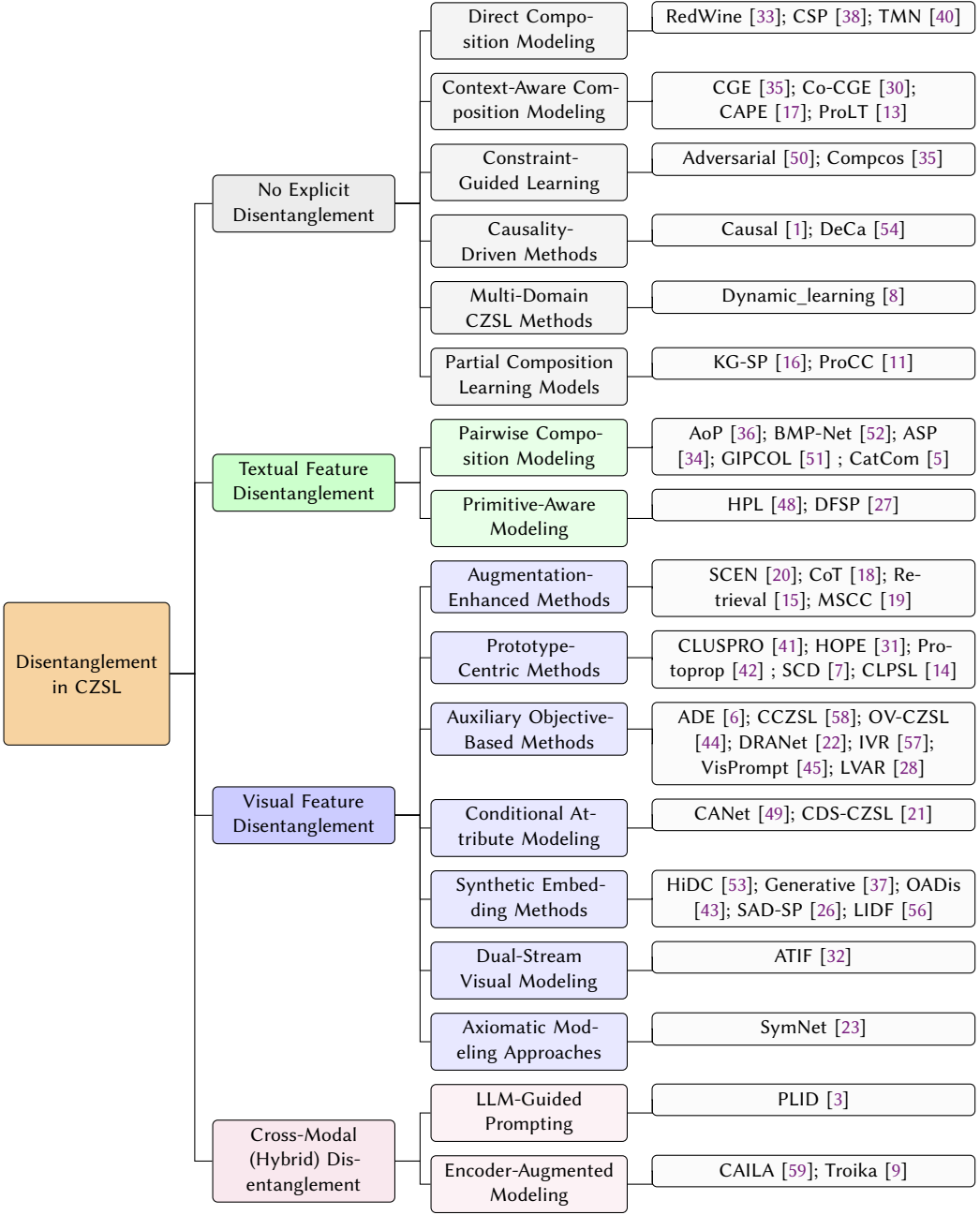


\section{Taxonomy: Disentanglement in CZSL}
\label{sec:taxonomy}

Building upon the principle of feature disentanglement introduced in Section \ref{sec:definitions}, we propose a comprehensive taxonomy for Compositional Zero-Shot Learning (CZSL) methods. This framework is grounded in the core challenges of CZSL and the strategies adopted to address them, offering a systematic framework for analysis. As illustrated in Fig. \ref{fig:taxonomy-diagram}, our taxonomy first categorizes methods by the form of disentanglement they pursue: no explicit disentanglement, textual, visual and cross-modal (hybrid) disentanglement. At a second level, methods are further clustered by their strategies for modeling attributes and addressing compositional challenges, including prototype-based modeling, synthetic embeddings, augmentation, causal inference, and constraint-driven formulations, among others. This hierarchical organization provides a precise and non-overlapping perspective on diverse design philosophies, enabling systematic comparison of underlying principles and revealing the evolving methodological focus of the field. The following subsections elaborate on each category of the taxonomy, examining their representative approaches, strengths, and limitations.

\subsection{No Explicit Disentanglement}
\label{sec:no-disentanglement}

\begin{figure}
    \centering
    \includegraphics[width=1.0\linewidth]{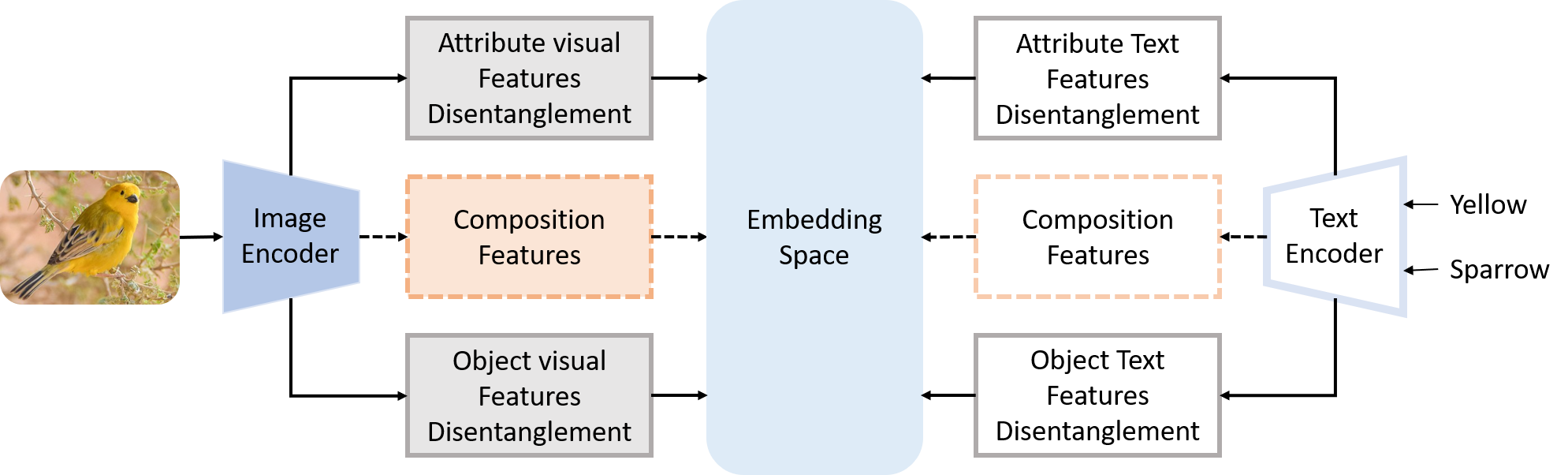}
    \caption{Architectural Overview: \textbf{No Explicit Disentanglement} bypasses primitive separation and models entire attribute-object compositions holistically through unified embeddings or simple fusion mechanisms (dotted boxes and dotted lines in the figure). \textbf{Textual Disentanglement} operates in the language space (gray boxes on the right), learning independent embeddings of attributes and objects that can be semantically composed at inference time. \textbf{Visual Disentanglement} focuses on the visual space (gray boxes on the left), isolating attributes and objects into structured, discriminative representations that can be systematically recombined into unseen compositions. \textbf{Cross-Modal Disentanglement} integrates both vision and language, disentangling primitives within each modality and aligning them in a shared embedding space to leverage complementary cues for more robust generalization.}
    \label{fig:methodology-diagram}
\end{figure}

The earliest approaches to Compositional Zero-Shot Learning (CZSL) did not explicitly disentangle attributes and objects, but instead treated each attribute-object composition as a single unit. In these models, both image features and textual labels were mapped directly into a shared embedding space, often through simple projection layers, as illustrated by the dashed orange rectangles in Fig. \ref{fig:methodology-diagram}. This design offered a conceptually straightforward formulation and established the first baselines for CZSL, while also benefiting from the use of pre-trained embeddings. However, by overlooking the internal structure of compositional concepts, these models were prone to overfitting to seen pairs and struggled to generalize to unseen compositions. Their limitations underscore the need for disentanglement-based approaches, but their simplicity makes them an important reference point for evaluating progress in the field.

Within this category, subsequent methods can be organized into subcategories that progressively address the absence of explicit primitive separation. \textbf{Direct Composition Modeling} represents the simplest line of work, treating compositions as indivisible units. \textbf{Context-Aware Composition Modeling} augments this baseline by incorporating contextual dependencies among states, objects, and their compositions. \textbf{Constraint-Guided Learning} introduces tailored objectives or structural constraints to regularize the embedding space. \textbf{Causality-Driven Methods} reinterpret CZSL through the lens of causal inference, aiming to move beyond spurious correlations and capture the underlying generative dependencies. \textbf{Multi-Domain CZSL Methods} extend robustness to variations in domains or visual styles. Finally, \textbf{Partial Composition Learning Models} are designed to learn effectively from partially-labeled data, addressing the challenge of incomplete supervision by using compositions where either the attribute or the object label is missing.   Together, these subcategories illustrate how the field sought to overcome the limitations of non-disentangled representations while retaining their conceptual simplicity.

\subsubsection{\textbf{Direct Composition Modeling}}

Direct Composition Modeling represents both the origins and a modern adaptation of CZSL research. Early works in this category established a simple yet foundational baseline: attribute-object pairs were constructed by directly combining primitive representations and treated as holistic units, without introducing additional disentanglement mechanisms. These approaches demonstrated that novel combinations could be approximated through straightforward embedding operations, providing a reference point for subsequent advances. More recently, the same principle of direct composition has been revisited in the context of large-scale vision-language models, where methods adapt pretrained multimodal knowledge to achieve stronger generalization. Despite the temporal gap, both early and modern approaches share the defining characteristic of modeling attribute-object pairs directly, making them natural members of this category. The following paragraph discusses representative approaches that illustrate this strategy.

RedWine \cite{misra2017red} pioneered CZSL by framing the fundamental challenge of balancing compositionality and contextuality, directly composing visual classifiers for unseen attribute-object pairs without relying on external linguistic supervision. Building on this foundation, Compositional Soft Prompting (CSP) \cite{csp2023} adapted vision-language model such as CLIP to the CZSL task, introducing a parameter-efficient prompting strategy where attributes and objects are treated as learnable tokens of a compositional vocabulary. Leveraging CLIP’s multimodal alignment, CSP improves generalization to unseen and compositions, demonstrating how prompt-based adaptation extends the principles of contextual composition modeling into the pretrained multimodal paradigm. Yet, despite these advances, direct composition methods still lack an explicit mechanism to capture the dependency structure among attributes, objects, and their combinations, a gap that motivated the rise of Context-Aware Composition Modeling.

\subsubsection{\textbf{Context-Aware Composition Modeling}}

A central challenge in Compositional Zero-Shot Learning (CZSL) is that the semantics and visual appearance of attributes and objects shift depending on their context. For example, ``wet'' alters a \textit{dog} differently than it alters a \textit{car}. Direct Composition Modeling methods overlook this variability by treating each pair independently, limiting their ability to transfer knowledge across compositions. Context-Aware Composition Modeling addresses this gap by explicitly capturing the dependency structures that link attributes, objects, and their combinations. By leveraging relational dependencies, such as the way ``old cycle'' and ``old car'' inform one another, these approaches enable more systematic generalization to unseen compositions. Representative methods under this subcategory are described below.

CGE \cite{naeem2021learning} formulates CZSL as a graph-structured problem, embedding states, objects, and compositions as interconnected nodes where message passing enforces semantic compatibility and supports generalization without relying on external resources. Its extension, Co-CGE \cite{mancini2021open}, adapts this idea to the open-world setting by incorporating feasibility scores to penalize unrealistic compositions, strengthening performance across evaluation protocols. In contrast to CGE and Co-CGE, which use graph neural networks to capture dependencies between primitives and their compositions, CAPE \cite{CAPE} uses self-attention mechanism to get the dependencies and transfer knowledge between compositions. This helps the model to learn context-aware compositions.

This broader view of contextuality enables the model to capture richer semantic correlations and achieve more robust generalization to unseen pairs. Yet, even with explicit context modeling, models may still struggle with ambiguous or closely related compositions, where fine-grained distinctions are easily blurred. To address these challenges, another line of work introduces tailored constraints into the learning process, designing specialized loss functions or regularization schemes that guide models toward more discriminative and consistent representations. This is the focus of the next section.

\subsubsection{\textbf{Constraint-Guided Learning}}

A distinct branch of CZSL research employs constraint-guided learning, motivated by the observation that standard objectives such as classification or metric learning often fail to capture fine-grained information of compositions or to prevent unfeasible compositions from being treated as valid. For example, closely related pairs like ``young tiger'' and ``young cat'' may be easily confused without additional regularization. To address this, constraint-guided approaches introduce specialized loss functions that encourage models to respect semantic distinctions and penalize infeasible compositions. By embedding such constraints directly into the training objectives, these methods improve robustness and generalization to unseen pairs.

Adversarial \cite{wei2019adversarial} enhances the embedding space for unseen attribute-object recognition through multi-scale feature integration and a quintuplet loss that incorporates semi-negative samples (e.g., old tiger vs. young tiger), improving discriminability among visually similar pairs. An adversarial framework further refines representations by generating positive and semi-negative samples, strengthening robustness. CompCos \cite{mancini2021open} extends this direction by introducing the Open-World CZSL setting, where all possible attribute-object combinations, including unfeasible ones, form the test space. To handle this, CompCos embeds images and compositions in a shared space and applies a feasibility-based margin loss to penalize unrealistic compositions (e.g., ripe dog), enabling more robust alignment across modalities. Together, these works illustrate how stronger objective functions and feasibility constraints advance both closed-world and open-world CZSL. While such constraints provide valuable structure, they still rely heavily on correlations within the training data. The next line of work, causality-driven approaches, pushes further by explicitly modeling attributes and objects as causal factors, aiming to overcome spurious correlations and achieve more stable generalization.

\subsubsection{\textbf{Causality-Driven Methods}}

While constraint-guided learning introduces useful structure into the embedding space, it remains limited by its reliance on correlations observed in training data. These correlations often break down when primitives are recombined in novel ways. Causality-driven methods take a more fundamental step by reframing attributes and objects as causal factors whose interventions generate images. By shifting from correlation to causal modeling, these approaches aim to learn representations that are stable across distribution shifts and more faithful to the underlying generative process of compositions, offering a principled pathway toward robust generalization in CZSL.

Within this line of work, Atzmon et al. \cite{atzmon2020causal} reformulate CZSL inference as identifying ``which intervention caused the image,'' introducing causal-inspired embeddings with conditional independence constraints to obtain stable representations under distribution shift. Building on this foundation, DECA \cite{DeCa} extends the formulation with a decomposable causal view that explicitly models the entangled relationship between attributes and objects by breaking causal effects into three learnable components, including a mediator for contextual interactions. This more fine-grained treatment of contextuality situates prior approaches as partial cases within its broader framework, underscoring the promise of causality as a guiding principle for compositional generalization. At the same time, however, causal formulations primarily target semantic entanglement and distributional bias, whereas real-world scenarios often present additional challenges arising from shifts in domains or visual styles. Addressing this broader spectrum of variability has motivated a distinct line of research, which we turn to next.

\subsubsection{\textbf{Multi-Domain CZSL Methods}}
A key limitation of most CZSL approaches is their assumption that unseen test compositions are drawn from the same distribution as training data, involving only unseen combinations of previously observed attributes and objects under consistent visual conditions. In practice, however, test images may also vary across visual domains, where differences in illumination, artistic rendering, or imaging modality substantially alter how concepts appear. To address this broader challenge, a new setting named Realistic Compositional Zero-Shot Learning (RCZSL) \cite{hu2024dynamic} was proposed that adds images from additional domains in the training and testing. As a result, their model is able to recognize images from multiple domains at inference time. This direction provides a more  real-world condition, though research is still at an early stage, with only one method proposed so far, underscoring the need for further exploration.

This direction highlights how variability in visual domains complicates compositional generalization, requiring models to adapt beyond the semantic space of attributes and objects. Yet, an equally fundamental obstacle arises from the supervision itself: existing datasets typically cover only a small subset of all possible attribute-object combinations. This sparsity leaves substantial portions of the compositional space unlabeled, making it necessary for models to infer compositions from incomplete supervision rather than depending on fully labeled training pairs.

\subsubsection{\textbf{Partial Composition Learning Models}}

Partial Compositional Learning Models address a fundamental practical challenge in CZSL: the assumption that every training sample is fully annotated with both attribute and object labels. In real-world datasets, this is rarely the case. Annotation is costly, incomplete, and often skewed, leaving many potential compositions unlabeled or only partially labeled. Such sparsity undermines the coverage of the compositional space and limits the scalability of conventional approaches. To overcome this limitation, partial compositional learning methods aim to exploit supervision that is available only for one primitive type (e.g., objects or attributes) while inferring or approximating the missing labels. By leveraging feasibility priors or compatibility modeling, these approaches seek to generalize compositional reasoning from incomplete supervision, thus moving CZSL closer to real-world deployment scenarios where full annotations cannot be guaranteed.

Knowledge-Guided Simple Primitives (KG-SP) \cite{kg-sp} introduces the Partial CZSL (pCZSL) setting, where only one primitive type (attribute or object) is labeled during training. It tackles this by training independent primitive classifiers and combining them with feasibility priors from external knowledge (e.g., ConceptNet \cite{speer2017conceptnet}), using pseudo-labeling to recover missing supervision. ProCC \cite{huo2024procc} further strengthens pCZSL by explicitly modeling cross-primitive dependencies through a Cross-Primitive Compatibility (CPC) module and adopting a progressive training paradigm, first learning objects, then conditioning attribute recognition, and finally refining both jointly. Unlike KG-SP, ProCC avoids reliance on external resources and achieves more robust generalization in both open-world and pCZSL, establishing a stronger data-driven framework for handling incomplete supervision. Despite these improvements, methods in this category continue to face inherent weaknesses that stem from the lack of explicit disentanglement, as discussed next.

\textbf{Limitations:} Methods without explicit disentanglement represent some of the earliest attempts at CZSL \cite{misra2017red} and are appealing in their conceptual simplicity, often relying on direct composition embeddings \cite{mancini2021open} or compatibility functions \cite{naeem2021learning} between attribute-object pairs. However, this simplicity comes at a cost. By treating compositions holistically without explicitly separating primitives, these approaches are unable to capture the distinct semantics of attributes and objects or to model their contextual variability across different pairings. As a result, they struggle to generalize when primitives are recombined in novel ways, limiting their effectiveness in scenarios where compositional diversity is high. These shortcomings motivate the next line of research, which turns to the textual domain, where attributes and objects are already represented as distinct words. Leveraging this natural disentanglement, methods aim to refine and align textual primitives in order to improve compatibility modeling and generalization.


\subsection{Textual Feature Disentanglement}
\label{sec:textual-disentanglement}

Text provides a natural starting point for compositional disentanglement, as attributes and objects are already represented as distinct words rather than entangled visual features. Yet naive reliance on pre-trained word embeddings proves insufficient for CZSL, as such embeddings rarely capture how primitives interact in context. This limitation motivates textual disentanglement: refining attribute and object embeddings and explicitly modeling their contextual compatibility \cite{asp, Gipcol}. By moving beyond shallow separability, these methods aim to construct textual representations that are both flexible and compositional, providing a stronger basis for generalization to unseen pairs.

While all textual disentanglement methods aim to capture contextuality for compositional generalization, they differ in how they balance the role of primitives and their compositions. \textbf{Pairwise Composition Modeling} focuses primarily on the attribute-object pair as the central unit of learning: primitives are initially disentangled, but once their interaction is modeled, the composition is treated holistically and its embedding is refined as a single entity. \textbf{Primitive-Aware Modeling}, on the other hand, preserves explicit representations of attributes and objects throughout training, ensuring that primitive-level embeddings remain robust while also constructing composition embeddings that interact with them. This dual emphasis allows the model to learn not only how primitives combine but also how they function independently, leading to richer contextualization and more reliable transfer to unseen pairs.

\subsubsection{\textbf{Pairwise Composition Modeling}}

A key limitation of naive textual disentanglement is that, while it separates attributes and objects as individual primitives, it often overlooks the contextual dependencies that emerge when they combine. ``Small'' means something very different when applied to a ``cat'' than to a ``plane,'' and ignoring such interactions leads to generic representations with weak generalization ability. Pairwise composition modeling addresses this by explicitly learning relationships between attributes and objects before recombining them into holistic composition embeddings. Within this paradigm, some methods employ self-attention to capture relational cues \cite{asp}, while others use graph-based reasoning to inject structural knowledge of compositional relations \cite{Gipcol}. By first disentangling primitives, then modeling their contextuality, and finally recombining them, pairwise approaches offer a principled mechanism for generalizing reliably from seen to unseen attribute-object pairs.

Blocked Message Passing Network (BMP-Net) \cite{BMP-Net} exemplifies pairwise composition modeling by addressing relation awareness, consistency, and disentanglement through a concept module that captures semantic relations via key-query message passing and a visual module that extracts primitive features. Its edge-blocking and branch-blocking mechanisms mitigate over-reliance on seen pairs and reduce co-occurrence bias, improving generalization to unseen compositions. Extending this line of work to the open-world setting, ASP \cite{asp} introduces contextual self-attention \cite{vaswani2017attention} between attribute and object embeddings and incorporates external knowledge from ConceptNet \cite{speer2017conceptnet} to filter unrealistic pairs, demonstrating how pairwise modeling can scale to vast, unconstrained label spaces. In contrast to these attention-based strategies, GIPCOL \cite{Gipcol} employs a graph neural network to inject compositional structure into soft prompts for CLIP, treating attributes and objects as nodes in a compositional graph and propagating information to model feasible relations. This graph-injected prompting provides a principled way to guide vision-language models toward capturing compositional rules and generalizing beyond seen pairs.

Pairwise modeling advances textual disentanglement by explicitly capturing how attributes and objects interact in context, yet it tends to prioritize the composition as a whole over the primitives themselves \cite{asp, Gipcol}. This raises the need for approaches that retain explicit representations of attributes and objects while still modeling their interactions. Primitive-aware methods respond to this challenge by integrating primitive-level concepts with pairwise relations, offering a more balanced and transferable foundation for compositional generalization. By preserving primitive identities alongside compositional structure, these methods reduce over-reliance on seen pair distributions and better support recognition of novel combinations.

\subsubsection{\textbf{Primitive-Aware Modeling}}

Pairwise modeling improves contextual reasoning by learning how attributes and objects interact, but it often diminishes the role of the individual primitives, treating the composition as the primary focus. This can hinder generalization to unseen combinations, since the distinctive information carried by individual attributes and objects is not fully preserved. Primitive-aware modeling addresses this gap by maintaining explicit representations of primitives alongside composition spaces, enabling the model to leverage both primitive-level concepts and pairwise interactions. By combining these two sources of information, such approaches reduce over-reliance on seen pair distributions, strengthen generalization to novel combinations, and offer greater interpretability in how primitives contribute to recognition. Representative methods are discussed below.

HPL \cite{HPL} introduces a hierarchical framework that models states, objects, and their compositions in separate embedding spaces, each serving as an expert that contributes complementary knowledge at inference. Leveraging CLIP, the approach learns hierarchical prompts with fixed word tokens to adapt pretrained vision-language knowledge for CZSL. By jointly capturing primitive and compositional representations, the framework mitigates bias toward seen pairs and resolves ambiguities, thereby improving generalization to unseen attribute-object combinations. Complementing the hierarchical expert framework, DFSP \cite{DFSP} introduces a soft prompt-based strategy to jointly represent states and objects while easing the domain gap between seen and unseen compositions. A learnable prompt structure encodes prefix, state, and object tokens to capture compositional semantics, while a decomposed fusion module disentangles state and object features from the language encoder and aligns them with image features in a cross-modal space. This design allows the model to associate primitives with images both individually and jointly, improving its ability to respond to unseen compositions and enhancing generalization.

\textbf{Limitations:} Although textual disentanglement benefits from the inherent separability of attributes and objects in the language space, it remains insufficient on its own. Word-level representations cannot capture the rich visual variability of attributes, which often change drastically in appearance depending on the object context \cite{DFSP, asp}. As a result, relying solely on textual space risks overlooking the entanglement that actually occurs in images. This limitation motivates the next line of research, which focuses on visual disentanglement, where attributes and objects must be explicitly separated within the visual representation to support more reliable and robust compositional reasoning.


\subsection{Visual Feature Disentanglement}
\label{sec:visual-disentanglement}
Visual disentanglement forms a central strand of research in CZSL because the visual domain presents primitives in highly entangled forms. Unlike the language domain, which provides primitives as distinct semantic concepts, visual features intertwine the two in ways that are difficult to separate. For example, the attribute ``ripe'' manifests differently on ``bananas'' and ``tomatoes,'' while objects such as ``dogs'' often co-occur with particular attributes like ``brown'' in training data, reinforcing spurious correlations. Without explicit separation, models risk conflating these dependencies and overfitting to seen pairs. Visual disentanglement addresses this challenge by learning distinct, transferable representations for attributes and objects directly from image features, thereby enabling primitives to be systematically recombined into novel compositions at test time.

In practice, visual disentanglement methods achieve this by explicitly separating attribute and object features within the visual feature space. An input image is first processed by an image encoder, after which attribute-specific and object-specific components are learned through dedicated mechanisms. These primitive representations are either projected into separate embedding spaces or composed and then mapped into a shared embedding space, as illustrated in Fig. \ref{fig:methodology-diagram}. On the textual side, attribute and object labels are simply encoded with a text encoder and mapped into the same space, without additional disentanglement operations. Training objectives typically enforce alignment between the visual and textual embeddings of the same primitive (e.g., encouraging the visual embedding of yellow to match its textual counterpart), ensuring that attributes and objects remain distinct yet composable. This structured representation enables models to flexibly recombine primitives and recognize unseen attribute-object pairs.

While all visual disentanglement-based methods aim to learn structured and transferable representations of primitives, they differ in the technical strategies used to enforce disentanglement and compositional generalization. We organize these approaches into seven subcategories, each reflecting a distinct mechanism for encouraging primitive separation and recombination:
\begin{itemize}
    \item \textbf{Augmentation-Enhanced Methods} exploit data augmentation such as feature perturbation, mixup, or composition resampling to increase robustness and promote disentanglement of primitives under varied contexts.
    \item \textbf{Prototype-Centric Methods} construct dedicated prototype representations for attributes and objects, ensuring that each primitive is anchored to a stable reference in the embedding space. Novel compositions can then be recognized by combining these prototypes.
    \item \textbf{Auxiliary Objective-Based Methods} incorporate additional training losses (e.g., orthogonality, mutual information, reconstruction) to encourage the separation of primitive features and improve their recomposability.
    \item \textbf{Conditional Attribute Modeling} treats attributes as conditional modifiers of object representations, disentangling them by explicitly modeling their effect on the visual features of objects.
    \item \textbf{Synthetic Embedding Methods} generate artificial or extrapolated embeddings for unseen compositions, effectively extending the embedding space beyond what is observed during training.
    \item \textbf{Dual-Stream Visual Modeling} employs two independent visual encoders whose outputs are fused, allowing the model to disentangle attributes and objects through complementary visual pathways.
    \item \textbf{Axiomatic Modeling Approaches} impose structural or logical constraints (e.g., commutativity, exclusivity, or causal assumptions) on how primitives are represented and combined, thereby enforcing principled disentanglement.
\end{itemize}

Collectively, these subcategories reflect the range of strategies explored for visual disentanglement in CZSL. They demonstrate how different design choices, spanning architectures, objectives, and constraints, contribute to learning primitives that can be effectively recombined. The following subsections examine each approach in greater depth, outlining their methodological foundations.

\subsubsection{\textbf{Augmentation-Enhanced Methods}}
Augmentation-enhanced methods address one of the key bottlenecks in CZSL: the limited and imbalanced coverage of attribute-object pairs in available datasets. Real-world data often features a long-tail distribution where many compositions are sparsely represented, while a few are oversampled, risking biased learning and limited generalization to unseen pairs. To mitigate this, augmentation techniques expand the effective training distribution by generating synthetic compositions, perturbing or blending existing samples, or retrieving complementary representations. This expanded variability exposes primitives across broader contexts, helping models disentangle their features more effectively and reducing reliance on spurious co-occurrences. As a result, augmentation not only alleviates data scarcity but also strengthens generalization to unseen compositions.

Considering the entanglement between state and object, SCEN \cite{SCEN} embeds the visual feature into a Siamese Contrastive Space to learn the primitives separately, facilitating the disentanglement between state and object. In addition, SCEN designs a State Transition Module (STM) to increase the diversity of training compositions, improving the robustness of the recognition model. The Composition Transformer (CoT) \cite{CoT} combines object-specific and attribute-specific experts with object-guided attention to model contextuality and introduces Minority Attribute Augmentation (MAA), which selectively mixes underrepresented attributes to address long-tailed distributions. Complementing these synthetic strategies, Retrieval \cite{Retrieval} leverages dynamically updated databases of primitive representations, retrieving and fusing similar instances to enrich input features with knowledge of related compositions. Together, these approaches demonstrate how augmentation through synthesis, balancing, or retrieval can reduce bias, increase diversity, and yield more discriminative embeddings for unseen attribute-object pairs. Beyond such data-driven enrichment, another important direction focuses on grounding primitives in stable, representative forms that can serve as reliable building blocks for composition, offering a complementary route to improving generalization.

\subsubsection{\textbf{Prototype-Centric Methods}}

While augmentation strategies enrich the training distribution and reduce spurious correlations, they operate only indirectly on the representation and do not provide a stable mechanism for separating primitives. Prototype-centric methods address this gap by building compositional reasoning around prototypical representations of attributes and objects. Instead of relying solely on raw embeddings or simple centroids, these approaches learn or retrieve prototypes that act as structured anchors for disentangling primitives and composing them into novel pairs. By explicitly modeling prototypes, they capture the diversity of primitive manifestations and establish consistent reference points for composition. This anchoring offers several benefits: it enforces independence at the primitive level, improves robustness by accounting for intra-class variability, and enhances generalization by treating prototypes as transferable building blocks for unseen compositions. Representative methods following this paradigm are discussed in detail below.

ProtoProp \cite{Protoprop} learns disentangled attribute and object prototypes and propagates them through a compositional graph, preserving useful correlations while avoiding spurious ones, without relying on external semantic resources. CLUSPRO \cite{CLUSPRO} addresses the limits of single-centroid representations by mining multiple diverse prototypes per primitive via clustering, combining contrastive and decorrelation losses to ensure both variability and independence; its non-parametric design makes it expressive yet efficient, and it achieves state-of-the-art performance. HOPE \cite{HOPE} takes a complementary direction, using a Modern Hopfield Network as associative memory to retrieve relevant prototypes and a Soft Mixture of Experts to fuse them with image features, enabling flexible and memory-driven compositional reasoning. Together, these approaches demonstrate how prototype-centric modeling can provide more robust and adaptable primitive representations for CZSL. Although prototype-centric approaches offer stable anchors for primitive representations, they rely heavily on the quality of the learned prototypes and provide limited guarantees of true disentanglement. To complement this, a parallel line of work introduces auxiliary objectives, which add regularization terms that explicitly enforce independence or compatibility, guiding the model beyond the primary classification task.

\subsubsection{\textbf{Auxiliary Objective-Based Methods}}

Architectural modifications and augmentation strategies improve compositional generalization, but they often lack explicit mechanisms to guarantee disentanglement or prevent knowledge degradation. Auxiliary objective-based methods address this gap by introducing specialized loss functions and regularization terms that directly enforce desired properties during training. By shaping the learning process through constraints such as primitive separation or knowledge retention, these objectives provide targeted guidance for overcoming key CZSL challenges: from disentangling highly entangled visual concepts to mitigating catastrophic forgetting and enabling open-vocabulary recognition. Representative methods following this paradigm are discussed below.


ADE \cite{ADE} employs Vision Transformer cross-attentions as disentanglers, pairing inputs that share a concept to separate attribute-exclusive and object-exclusive representations, while an Earth Mover’s Distance regularization enforces consistency and reduces leakage across concepts. Extending beyond standard settings, CCZSL \cite{zhang2024continualCZSL} introduces continual CZSL, where primitives expand incrementally; it mitigates contextual shifts and catastrophic forgetting through super-primitive abstraction and dual knowledge distillation. In parallel, OV-CZSL \cite{OV-CZSL} proposes an open-vocabulary setting that requires recognizing unseen attributes and objects in addition to their unseen combinations at inference time. Leveraging pre-trained language embeddings and a Neighborhood Expansion Loss, OV-CZSL aligns visual features with semantically related linguistic concepts, enabling generalization beyond fixed vocabularies. Together, these approaches highlight how auxiliary objectives can support both disentanglement and generalization in challenging new CZSL settings.

While auxiliary objectives expand the representational space and improve generalization to unseen primitives, they often treat attributes in isolation, without explicitly modeling how their meaning changes with different objects. Conditional attribute modeling addresses this limitation by learning attribute representations that are dynamically conditioned on object context, enabling more precise and context-sensitive compositional recognition.

\subsubsection{\textbf{Conditional Attribute Modeling}}

Many CZSL approaches, including those enhanced with auxiliary objectives, continue to assume that attributes can be represented as fixed, object-agnostic embeddings. In practice, however, attributes vary dramatically across contexts: ``ripe'' manifests differently for apples than for bananas, and ``small'' conveys different semantics for animals versus vehicles. Treating attributes as context-independent entities leads to generic representations that overlook this variability, limiting compositional generalization. Conditional Attribute Modeling addresses this issue by adapting attribute embeddings to the specific object or context in which they occur, producing more flexible and discriminative representations. By moving beyond static attribute learning and explicitly conditioning on objects, images, or contextual cues, these methods strengthen the ability to distinguish fine-grained attribute-object pairs. Representative approaches are discussed below.


CANet \cite{CANet} introduces a dual learner framework, where a hyper learner extracts prior knowledge from object and image context to guide a base learner in generating conditional attribute embeddings. This design captures how attributes vary across objects (e.g., wet apple vs. wet cat), improving generalization to unseen pairs. Building on this, CDS-CZSL \cite{CDS-CZSL} further models attribute specificity, recognizing that some attributes (e.g., sliced) are more discriminative than others (e.g., red). By conditioning on objects, images, and attribute informativeness, CDS-CZSL provides a principled mechanism for prioritizing and filtering attribute-object pairs, advancing conditional modeling beyond simple adaptation.


Although conditional attribute modeling provides richer, context-sensitive representations, its effectiveness remains tied to the diversity of contexts observed during training. When attributes or objects appear in rare or entirely unseen combinations, conditioning alone may fail to provide reliable and accurate embeddings. To overcome this limitation, a complementary line of research explores methods, which explicitly generate representations for unseen compositions. By synthesizing attribute-object embeddings, these methods proactively address compositional novelty, extending the representational space beyond the training set.

\subsubsection{\textbf{Synthetic Embedding Methods}}

A persistent challenge in CZSL is that many methods, even those adapting attributes to context, remain limited by the combinations observed during training. When entirely new attribute-object pairs arise, the model lacks guidance for how such concepts should be represented. To address this, synthetic embedding methods generate novel compositions directly in the embedding space, allowing models to anticipate unseen pairs without requiring real images. This strategy alleviates data sparsity, enriches training diversity, and offers a more proactive way to model fine-grained contextuality. By explicitly constructing embeddings for unseen compositions, these approaches extend the effective training space and improve robustness to compositional novelty. Representative methods are discussed below.

HiDC \cite{HiDC} introduces a hierarchical decomposition-recomposition framework that generates synthetic visual, textual, and hybrid embeddings, refined through semi-positive concepts, to capture fine-grained contextual correlations. Nan et al. \cite{nan2019recognizing} instead adopt an encoder-decoder design that unifies visual and linguistic features in a latent space, preserving primitive properties while modeling their interdependence for robust unseen recognition. In contrast, OADis \cite{OADis} focuses purely on the visual domain, disentangling attributes and objects via contrastive pairings (e.g., peeled apple vs. sliced apple) and recombining them to hallucinate unseen pairs. Together, these generative strategies highlight the value of creating synthetic embeddings, which can be either multimodal or visual-only, for improving compositional generalization. Although synthesizing embeddings expands the training space and improves robustness to unseen pairs, these methods still operate within a single representational stream, which can limit their ability to capture complementary visual cues. To address this, another line of work employs dual encoders, learning parallel streams of visual features that are later fused to strengthen disentanglement and compositional reasoning.

\subsubsection{\textbf{Dual-Stream Visual Modeling}}
Single-stream models, even when enriched with synthetic embeddings, remain limited in their ability to capture the full variability of compositional concepts. Different visual encoders excel at different aspects: conventional backbones provide detailed local features, while large-scale vision-language encoders contribute semantically aligned representations. Dual-stream modeling seeks to harness these complementary strengths by learning parallel visual encoders and fusing their outputs, thereby offering a richer basis for disentangling attributes and objects. This direction remains largely unexplored despite its promise, as only one method has been proposed to date, which highlights both its novelty and its potential. ATIF \cite{min2024adaptive} introduces a dual-stream framework that fuses traditional visual encoders with VLM image encoders to improve compositional generalization. An Adaptive Fusion Module (AFM) learns dynamic weights to combine similarity scores from both streams, while a Multi-Attribute Object Module (MAOM) enriches primitive representations by incorporating multiple attribute-object pairs. Built on top of a VLM text encoder, this design balances fine-grained visual features with semantic alignment, yielding robust recognition of unseen compositions.


Despite the potential advantages of combining the complementary strengths of traditional visual encoders and vision-language model encoders, dual-stream modeling also presents certain drawbacks. The representations produced by the two streams may differ significantly, making fusion complex and sometimes inconsistent. Without careful balancing, such discrepancies could introduce noise or reduce the stability of learned embeddings. Since only one method has explored this design so far, further investigation is required to validate its effectiveness, refine fusion mechanisms, and assess whether the benefits of integrating heterogeneous encoders outweigh the risks of representational misalignment. While dual-stream modeling explores architectural diversity to capture complementary representations, it does not explicitly define the principles by which attributes and objects should interact. Axiomatic approaches take a different route, introducing formal constraints or rules that guide composition directly, offering a more principled way to enforce disentanglement and compatibility.

\subsubsection{\textbf{Axiomatic Modeling Approaches}}
Most CZSL approaches rely on data-driven supervision or learned embeddings, which, while effective, provide no guarantees about the logical consistency of attribute-object transformations. This gap motivates axiomatic modeling, where formal principles define how primitives should interact. Drawing on structures such as group theory, these methods embed axioms like symmetry and invertibility into the learning process, ensuring that composition and decomposition behave in rational and interpretable ways. By grounding compositional reasoning in formal rules rather than heuristics, this line of work offers a principled alternative to purely empirical strategies. Although still in its infancy, with only one method proposed so far, it highlights an important direction for building consistency into CZSL.

SymNet \cite{li2020symmetry} introduces an axiom-driven framework for CZSL that models attribute-object interactions as formal transformations inspired by group theory. It enforces principles such as symmetry, closure, and invertibility through Coupling and Decoupling Networks for adding or removing attributes. Attribute recognition is guided by a Relative Moving Distance (RMD) criterion, which focuses on attribute-induced changes rather than raw appearance. Extended to multi-attribute cases, SymNet provides structured and interpretable embeddings, offering a principled alternative to purely data-driven methods. \\


Visual disentanglement offers several key advantages for compositional zero-shot learning. By explicitly separating attributes and objects into independent yet recombinable representations, these methods yield more discriminative and robust embeddings that support systematic generalization to unseen combinations \cite{SCEN, Retrieval, CLUSPRO, CDS-CZSL, OADis}. This decomposition also reduces the learning burden: rather than treating every attribute-object pair as a separate class, the model needs only to learn a smaller set of primitives and their interactions, improving both data efficiency and scalability. Another advantage is interpretability, as disentangled representations allow us to inspect how individual attributes and objects contribute to predictions. Crucially, unlike the textual domain, where primitives such as words are inherently separable, visual features entangle attributes and objects within raw image representations. Explicit disentanglement is therefore essential to make compositional reasoning tractable. Finally, in open-world scenarios where the space of possible compositions grows combinatorially, visual disentanglement provides a principled means of managing this complexity by enabling a limited set of primitives to be recombined into an effectively unbounded number of unseen pairs. Together, these strengths position visual disentanglement as a powerful paradigm for CZSL. However, its advantages come with trade-offs that raise several challenges.

\textbf{Limitations:} Despite its promise, visual disentanglement also faces notable limitations. Forcing strict separation between attributes and objects can oversimplify their natural dependencies, discarding contextual cues that may aid recognition. Implementations often require additional architectural modules, auxiliary losses, or specialized training procedures, which increase both complexity and computational cost. Achieving clean separation is also difficult in practice, since attributes and objects are not always easily distinguishable in visual space, particularly for attributes that are subtle, localized, or strongly context-dependent (e.g., ripe or broken). These challenges can result in noisy or unstable embeddings. Finally, while disentanglement is motivated by scalability, its effectiveness depends heavily on the quality of supervision and the diversity of training data. Without broad coverage, the learned primitives may fail to generalize to rare or fine-grained combinations. These limitations highlight the need for approaches that move beyond vision alone, motivating hybrid strategies that leverage both vision and text modalities to enforce disentanglement and improve compositional generalization.

\subsection{Cross-Modal (Hybrid) Disentanglement}
\label{sec:cross-modal-disentanglement}

Neither visual nor textual disentanglement alone is sufficient for robust compositional generalization. Visual features capture appearance-level variability but are deeply entangled and context-dependent, while textual labels offer discrete semantics yet lack grounding in perception. Relying on either modality in isolation therefore leads to incomplete representations: visual disentanglement struggles to handle fine-grained variability, and textual disentanglement misses the nuances of visual realization. Cross-modal disentanglement addresses this by simultaneously separating primitives in both modalities and aligning them within a shared embedding space. In doing so, it grounds linguistic semantics in visual perception while preserving the flexibility needed to capture diverse attribute-object instantiations, offering a more comprehensive basis for reasoning over unseen compositions.

Within this category, existing methods diverge in how they integrate visual and textual spaces. \textbf{LLM-Guided Prompting} draws on the semantic richness of large language models to craft or adapt prompts that encode compositional structure, guiding the alignment of visual features with contextualized textual embeddings. \textbf{Encoder-Augmented Modeling}, in contrast, reinforces multimodal encoders with specialized modules or auxiliary objectives that disentangle and align primitives across both modalities. These two directions reflect complementary strategies: one emphasizing semantic guidance from powerful language models, the other strengthening the multimodal encoder’s internal representations for joint disentanglement.

\subsubsection{\textbf{LLM-Guided Prompting}}

Prompting is a crucial mechanism for adapting vision-language models such as CLIP to compositional tasks, yet existing techniques \cite{DFSP, HPL} often rely on rigid templates or limited learnable tokens that fail to capture fine-grained variability and contextual richness. This limitation motivates the use of Large Language Models (LLMs), which can generate semantically diverse and context-sensitive descriptions that serve as more informative class representations. By grounding prompts in richer linguistic semantics, LLM-guided approaches offer a way to improve primitive disentanglement and enhance generalization to unseen attribute-object combinations. Although still at an early stage, initial work demonstrates the promise of this direction, suggesting that LLM-informed prompting could become a powerful tool for cross-modal CZSL.

{\footnotesize
\begin{longtable}{clcccccc}

    \caption{Summary of published CZSL papers. MM, TPAMI, TMM, PR stands for IEEE Multimedia, IEEE Transactions of Patern Analysis and Machine Intelligence and IEEE Transactions on Multimedia, Pattern Recognition respectively. Un, CW, OW, OV Par. stands for Unseen, Close-world, Open-world, Open Vocabulary and Partial respectively. FT, W2V, MIT and UT stands for Fasttext, Word2vec, MIT-States and UT-Zappos respectively.}
    \label{tab:summary} \\
    \toprule
    & \textbf{Model} & \textbf{Venue} & \textbf{Year} & \textbf{Test setting} & \textbf{Vision Encoder} & \textbf{Text Encoder} & \textbf{Datasets} \\
    \midrule
    \endfirsthead 

    \toprule 
    & \textbf{Model} & \textbf{Venue} & \textbf{Year} & \textbf{Test setting} & \textbf{Vision Encoder} & \textbf{Text Encoder} & \textbf{Datasets} \\
    \midrule 
    \endhead 


    {\multirow{20}{0.3cm}{
    \rotatebox[origin=c]{90}{\textbf{No Explicit Disentanglement}}
    }} & \multicolumn{7}{c}{\textbf{Direct Composition Modeling}} \\
    & RedWine \cite{misra2017red} & CVPR & 2017 & Un & VGG & - & MIT, StanfordVRD \\
    & CSP \cite{csp2023} & ICLR & 2023 & CW, OW & Clip & Clip & MIT, UT, CGQA \\
    & TMN \cite{purushwalkam2019task} & ICCV & 2019 & CW & ResNet & Glove & MIT, UT \\
    \cdashline{2-8} 
    & \multicolumn{7}{c}{\textbf{Context-Aware Composition Modeling}} \\
    & CGE \cite{naeem2021learning} & CVPR & 2021 & CW & ResNet & FT, W2V & MIT, UT, CGQA \\
     & Co-CGE \cite{mancini2022learning} & TPAMI & 2022 & CW, OW & ResNet & FT, W2V & MIT, UT, CGQA \\
    & CAPE \cite{CAPE} & WACV & 2023 & CW & ResNet & FT, W2V & MIT, UT, CGQA \\
    & ProLT \cite{ProLT} & AAAI & 2024 & CW & Clip & Clip & MIT, UT, CGQA \\
    \cdashline{2-8} 
    & \multicolumn{7}{c}{\textbf{Constraint-Guided Learning}} \\
    & Adversarial \cite{wei2019adversarial} \\
    & Compcos \cite{mancini2021open} & CVPR & 2021 & CW, OW & ResNet & FT, W2V & MIT, UT \\
    \cdashline{2-8} 
    & \multicolumn{7}{c}{\textbf{Causality-Driven Methods}} \\    
    & Causal \cite{atzmon2020causal} & NeurIPS & 2020 & Un, CW & ResNet & & UT, AO-CLEVr \\
    & DeCa \cite{DeCa} & TMM & 2022 & CW & ResNet & FT, W2V & MIT, UT \\
    \cdashline{2-8} 
    & \multicolumn{7}{c}{\textbf{Multi-Domain CZSL Methods}} \\
    & Dynamic \cite{hu2024dynamic} & AAAI & 2024 & CW & Clip & Glove& MIT, UT, CGQA \\
    \cdashline{2-8} 
    & \multicolumn{7}{c}{\textbf{Partial Composition Learning Models}} \\
    & KG-SP \cite{kg-sp} & CVPR & 2022 & OW, Par. & ResNet & None & MIT, UT, CGQA \\
    & ProCC \cite{huo2024procc} & AAAI & 2024 & OW, Par. & ResNet & - & MIT, UT, CGQA \\
    \midrule
    {\multirow{9}{0.3cm}{
    \rotatebox[origin=c]{90}{\textbf{Textual Disentanglement}}
    }}
    & \multicolumn{7}{c}{\textbf{Pairwise Composition Modeling}} \\
    & AoP \cite{nagarajan2018attributes} & ECCV & 2018 & Un, CW & ResNet & Glove & MIT, UT \\
    & BMP-Net \cite{BMP-Net} & MM & 2021 & CW & ResNet & - & MIT, UT \\
    & ASP \cite{asp} & DICTA & 2024 & OW & ResNet & W2V & MIT, UT, CGQA \\
    & GIPCOL \cite{Gipcol} & WACV & 2024 & CW, OW & Clip & Clip & MIT, UT, CGQA \\
    & CatCom \cite{Catcom} & ECCV & 2024 & CW, OW & ResNet & Glove & MIT, UT, CGQA \\
    \cdashline{2-8} 
    & \multicolumn{7}{c}{\textbf{Primitive-Aware Modeling}} \\
    & HPL \cite{HPL} & IJCAI & 2023 & CW, OW & Clip & Clip & MIT, UT, CGQA \\
    & DFSP \cite{DFSP} & CVPR & 2023 & CW, OW & Clip & Clip & MIT, UT, CGQA \\
    \midrule
    & \multicolumn{7}{c}{\textbf{Augmentation-Enhanced Methods}} \\
    & SCEN \cite{SCEN}& CVPR & 2022 & CW & ResNet & - & MIT, UT, CGQA \\
    & CoT \cite{CoT} & ICCV & 2023 & CW & ViT & Glove & MIT, Cgqa, VAW \\
    & Retrieval \cite{Retrieval} & AAAI & 2024 & CW, OW & Clip & Clip & MIT, UT, CGQA \\
    \cdashline{2-8} 
    {\multirow{25}{0.3cm}{
    \rotatebox[origin=c]{90}{\textbf{Visual Disentanglement}}
    }}
    & \multicolumn{7}{c}{\textbf{Prototype-Centric Methods}} \\
    & ProtoProp \cite{Protoprop} & NeurIPS & 2021 & CW & ResNet & - & AO-Clevr, MIT, UT \\
    & CLUSPRO \cite{CLUSPRO} & ICLR & 2025 & CW, OW & Clip & Clip & MIT, UT, CGQA \\
    & HOPE \cite{HOPE} & WACV & 2025 & CW, OW & Clip & Clip & MIT, UT, CGQA \\
    & SCD \cite{SCD} & AAAI & 2023 & CW & ResNet & - & UT, CGQA \\
    & CLPSL \cite{CLPSL} & TMM & 2025 & CW, OW & Clip & Clip & Clothing, UT, CGQA \\
    \cdashline{2-8} 
    & \multicolumn{7}{c}{\textbf{Auxiliary Objective-Based Methods}} \\
    & ADE \cite{ADE} & CPVR & 2023 & CW, OW & ViT & W2V & Clothing, UT, CGQA \\
    & CCZSL \cite{zhang2024continualCZSL} & IJCAI & 2024 & CCZSL & ResNet & - & UT, CGQA \\
    & OV-CZSL \cite{OV-CZSL} & CVPR & 2024 & OV & ResNet & Bert & MIT, CGQA, VAW \\
    & DRANet \cite{DRANet} & ICCV & 2023 & OW & ResNet & - & MIT, UT, CGQA \\
    & IVR \cite{IVR} & ECCV & 2022 & CW & ResNet & - & MIT, UT \\
    & VisPrompt \cite{VisPrompt} & PR & 2025 & CW, OW & Clip & Clip & MIT, UT, CGQA \\
    & LVAR \cite{ma2024lvar} & TCSVT & 2024 & CW & ResNet & Glove, W2V & MIT, UT, CGQA, VAW \\
    \cdashline{2-8} 
    & \multicolumn{7}{c}{\textbf{Conditional Attribute Modeling}} \\
    & CANet \cite{CANet} & CPVR & 2023 & CW & ResNet & FT, W2V & MIT, UT, CGQA \\
    & CDS-CZSL \cite{CDS-CZSL} & CVPR & 2024 & CW, OW & Clip & Clip & MIT, UT, CGQA \\
    & \multicolumn{7}{c}{\textbf{Synthetic Embedding Methods}} \\
    & HiDC \cite{HiDC} & CVPR & 2020 & Un, CW & ResNet & - & MIT, UT \\
    & Generative \cite{nan2019recognizing} & AAAI & 2019 & Un & ResNet & Glove & MIT, UT \\
    & OADis \cite{OADis} & CVPR & 2022 & CW & ResNet & Glove & MIT, UT, VAW \\
    & SAD-SP \cite{SAD-SP} & TPAMI & 2023 & OW & ResNet & - & MIT, UT, CGQA \\
    & LIDF \cite{LIDF} & TPAMI & 2024 & CW, OW & ViT & - & UT, CGQA \\
    \cdashline{2-8} 
    & \multicolumn{7}{c}{\textbf{Dual-Stream Visual Modeling}} \\
    & ATIF \cite{min2024adaptive} & TMM & 2024 & CW, OW & Clip, ViT & Clip & Clothing, UT, CGQA \\
    \cdashline{2-8} 
    & \multicolumn{7}{c}{\textbf{Axiomatic Modeling Approaches}} \\
    & SymNet \cite{li2021learning} & TPAMI & 2021 & Un, CW & ResNet & Glove & MIT, UT \\
    \midrule
    {\multirow{6}{0.3cm}{
    \rotatebox[origin=c]{90}{\textbf{Cross-Modal}}
    }} 
    & \multicolumn{7}{c}{\textbf{LLM-Guided Prompting}} \\
    & PLID \cite{PLID} & ECCV & 2024 & CW, OW & Clip & Clip & MIT, UT, CGQA \\
    \\ \cdashline{2-8} 
    & \multicolumn{7}{c}{\textbf{Encoder-Augmented Modeling}} \\
    & CAILA \cite{caila} & WACV & 2024 & CW, OW & Clip & Clip & MIT, UT, CGQA, VAW \\
    & Troika \cite{huang2024troika} & CVPR & 2024 & CW, OW & Clip & Clip & MIT, UT, CGQA \\ 
    
    
    
    \bottomrule

\end{longtable}
}

PLID \cite{PLID} enhances compositional zero-shot learning by addressing the limitations of prior prompting methods. Instead of using fixed templates, it leverages Large Language Models (LLMs) to create linguistically rich class distributions, capturing a greater variety of nuances and contextual cues. Additionally, a Visual-Language Primitive Decomposition (VLPD) module separates attribute and object features, improving robustness and reducing entanglement. By reframing classification as prompting over these distributions, PLID creates more expressive embeddings and achieves stronger generalization for unseen attribute-object pairs. While LLM-guided prompting enriches compositional representations through more expressive textual cues, it still operates primarily at the input level and leaves the underlying encoders unchanged. This raises the question of whether deeper improvements can be achieved by refining the internal structure of encoders themselves. Encoder-augmented approaches pursue this goal by directly enhancing how vision-language models disentangle and align attributes and objects within their encoders.

\subsubsection{\textbf{Encoder-Augmented Modeling}}
Prompt-based methods adapt vision-language models at the surface level, but leave the internal encoders unchanged, limiting their ability to disentangle primitives or capture fine-grained contextuality. Encoder-augmented modeling addresses this limitation by refining the internal structure of VLMs. By explicitly modeling attributes, objects, and their interactions within the encoder, these approaches leverage the rich multimodal knowledge of large pre-trained models while targeting their weaknesses in compositional generalization. This deeper intervention provides greater control over how primitives are separated, how contextuality is represented, and how embeddings align across modalities, leading to more robust transfer to unseen attribute-object pairs. Representative methods in this direction are discussed below.

Concept-Aware Intra-Layer Adapters (CAILA) \cite{caila} advances CZSL by moving beyond prompt tuning and external adapters that treat CLIP as a fixed black box. Instead, it inserts lightweight, concept-aware adapters directly within each layer of CLIP’s encoders, enabling the extraction of features specific to attributes, objects, and their compositions throughout the model. These adapters are combined via a Mixture-of-Adapters mechanism, which fuses knowledge from multiple concept-specific pathways to yield more discriminative and generalizable embeddings. To further enhance compositional learning, CAILA introduces Primitive Concept Shift, a data augmentation strategy that synthesizes novel vision embeddings by recombining attribute and object features across images. By explicitly disentangling and modeling primitives while retaining computational efficiency, CAILA demonstrates the importance of intra-layer concept specialization for improving compositional generalization in both closed-world and open-world settings.

Similar to CAILA, which extracts concept-specific features for states, objects, and compositions through intra-layer adapters, Troika \cite{huang2024troika} also adopts a primitive-aware perspective by explicitly modeling states and objects alongside compositions. Building on this idea, Troika introduces a Multi-Path paradigm that establishes three parallel branches for state, object, and composition modeling, aligning branch-specific prompts with decomposed visual features. By integrating predictions across these branches, the framework fully exploits pre-trained VLM knowledge while reducing over-reliance on seen compositions. Furthermore, a Cross-Modal Traction module dynamically adjusts prompt representations toward the visual content, mitigating modality bias and enhancing adaptability. Together, these innovations make Troika a flexible and generalizable framework for compositional reasoning in CZSL. Despite the progress achieved by encoder-augmented approaches such as Troika, cross-modal disentanglement as a whole still faces fundamental challenges that limit its scalability and robustness, which we discuss next.

\textbf{Limitations:} Cross-modal disentanglement has emerged as one of the most powerful directions in CZSL, as it combines the fine-grained grounding of visual features with the structured semantics of language. This dual perspective enables richer compositional reasoning and explains why much of the latest research has shifted toward hybrid modeling. Yet several challenges remain unresolved. Foremost among these is architectural and computational complexity: aligning two modalities often demands additional modules such as adapters, alignment networks, or specialized prompting strategies, increasing training overhead and design difficulty. More fundamentally, consistent cross-modal grounding remains elusive, since visual primitives vary widely in appearance while textual primitives remain comparatively stable. These limitations highlight that, despite impressive progress, cross-modal disentanglement is still an open frontier, and it continues to attract active research aimed at building more robust and generalizable compositional systems. The key details of all existing CZSL models, including their datasets, encoders, and taxonomy categories, are summarized in Table \ref{tab:summary}.

\section{CZSL Datasets}
\label{sec:datasets}
\subsection{MIT-States}
Used extensively in Compositional Zero-Shot Learning research, the MIT-States dataset \cite{mit-states} contains 53,753 images annotated with 115 distinct states and 245 different objects. This results in a total of 1,962 unique attribute-object compositions. For the standard CZSL setup, 1,262 of these compositions are designated as ``seen'' for training, leaving 700 as ``unseen'' for validation and testing in the closed-world setting. MIT-States has been a prominent dataset in CZSL since its introduction; however, previous research \cite{atzmon2020causal} has identified erroneous data within the dataset. Table \ref{tab:dataset} presents a detailed statistical overview of the dataset, and Table \ref{tab:images} includes representative images.

\begin{table}
  \caption{Datasets used in Compositional Zero Shot Learning with statistics. State s, Object o, Seen Composition $c_{s}$, Unseen Compositions $c_{u}$ and Images $c_{i}$.}
  \label{tab:dataset}
  \begin{tabular}{lcc|cc|ccc|ccc}
    \toprule
     &  &  & \multicolumn{2}{c|}{Training}  & \multicolumn{3}{c|}{Validation} & \multicolumn{3}{c}{Test} \\
     Dataset & s & o & $c_{s}$ & $c_{i}$ & $c_{s}$ & $c_{u}$ & $c_{i}$ & $c_{s}$ & $c_{u}$ & $c_{i}$ \\
     \midrule
     MIT-States \cite{mit-states} & 115 & 245 & 1262 & 30338 & 300 & 300 & 10420 & 400 & 400 & 12995 \\
     UT-Zappos \cite{ut-zappos} & 16 & 12 & 83 & 22998 & 15 & 15 & 3214 & 18 & 18 & 3000 \\
    CGQA \cite{naeem2021learning} & 413 & 674 & 5592 &  26920 & 1252 & 1040 & 7280 & 888 & 923 & 5098 \\
    VAW-CZSL \cite{vaw-czsl} & 440 & 541 & 11175 & 72203 & 2121 & 2322 & 9524 & 2449 & 2470 & 10856 \\
    Clothing16K & 9 & 8 & 18 & 7242 & 10 & 10 & 5515 & 9 & 8 & 3413 \\
    MAD & 158 & 309 & 5630 & 81371 & 1000 & 1000 & 27104 & 1400 & 1400 & 42,013 \\
    
  \bottomrule
\end{tabular}
\end{table}

\subsection{UT-Zappos}
The UT-Zappos dataset \cite{ut-zappos} is a shoe dataset characterized by 16 attribute concepts primarily related to material (e.g., ``Canvas'') and 12 object concepts representing shoe types (e.g., ``Boots''). The dataset is divided into training, validation, and test sets. The training set contains 23,000 images representing 83 compositions. The validation set consists of 3,000 images covering 15 seen and 15 unseen compositions, while the test set also has 3,000 images, split into 18 seen and 18 unseen compositions, as detailed in Table \ref{tab:dataset}. The smaller scale of the UT-Zappos dataset renders it computationally efficient for model validation, facilitating quicker experimental iterations. Conversely, its specialized content, exclusively featuring various shoe categories, poses a limitation in assessing the true generalization prowess of a model. Example images from the UT-Zappos dataset are shown in Table \ref{tab:images}.

\subsection{CGQA}
The CGQA dataset \cite{naeem2021learning} is derived from the Stanford GQA dataset \cite{hudson2019gqa}. It comprises 413 states and 674 objects. The dataset is divided into training, validation, and testing sets. The training set contains 27,000 images and includes 5,592 compositions. The validation set consists of 7,000 images with 1,040 unseen compositions, and the test set has 5,000 images covering 923 unseen compositions. CGQA represents a contemporary dataset that offers a strong benchmark for evaluating model generalization capabilities and accuracy. Currently, it stands as the premier dataset widely adopted for model assessment. Detailed statistics for the CGQA dataset are presented in Table \ref{tab:dataset}, and example images are shown in Table \ref{tab:images}.

\begin{table}[t]
\caption{Images and their compositions from the MIT-States, UT-Zappos, and CGQA datasets. MIT-States and CGQA are general-purpose datasets, whereas UT-Zappos is specialized for shoe types.}
\begin{center}
\begin{tabular}{cccc}
\toprule
\multicolumn{4}{c}{\textbf{MIT-States}} \\
\midrule
 \includegraphics[width=0.19\linewidth, height=2.2cm]{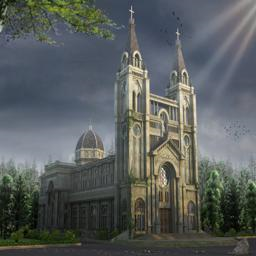} & \includegraphics[width=0.19\linewidth, height=2.2cm]{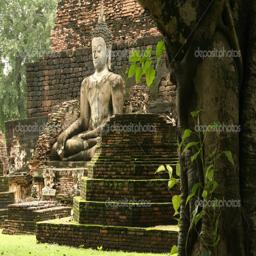} & \includegraphics[width=0.19\linewidth, height=2.2cm]{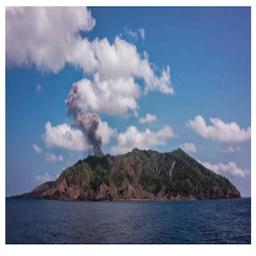} & \includegraphics[width=0.19\linewidth, height=2.2cm]{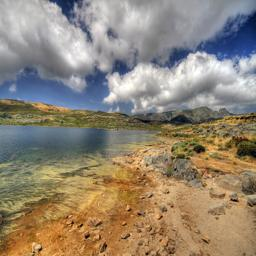}  \\
 Ancient Church & Ancient Jungle & Barren Island & Barren Lake \\
\midrule
\multicolumn{4}{c}{\textbf{CGQA}} \\
\midrule
\includegraphics[width=0.19\linewidth, height=2.2cm]{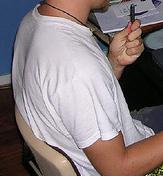} & \includegraphics[width=0.19\linewidth, height=2.2cm]{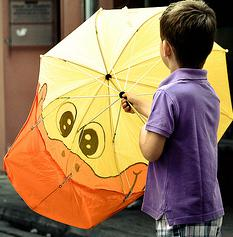} & \includegraphics[width=0.19\linewidth, height=2.2cm]{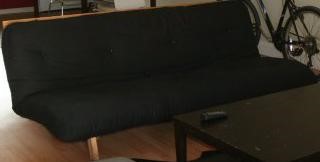}  & \includegraphics[width=0.19\linewidth, height=2.2cm]{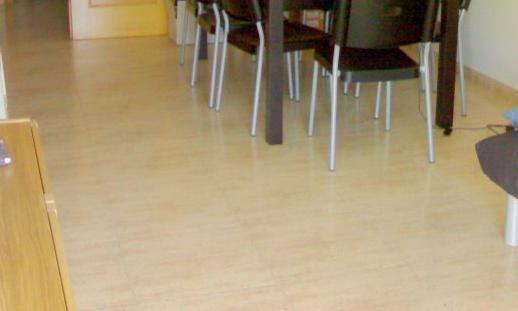} \\

 White Shirt & Yellow Umbrella & Black Couch & Tan Floor \\
\midrule
\multicolumn{4}{c}{\textbf{UT-Zappos}} \\
\midrule
 \includegraphics[width=0.19\linewidth, height=2.2cm]{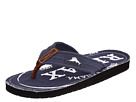} & \includegraphics[width=0.19\linewidth, height=2.2cm]{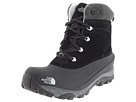} & \includegraphics[width=0.19\linewidth, height=2.2cm]{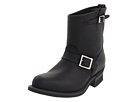}  & \includegraphics[width=0.19\linewidth, height=2.2cm]{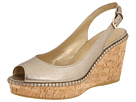} \\

 Canvas Sandal & Fleece Boots & Leather Boots & Nappa Sandals \\
\bottomrule
\end{tabular}
\end{center}

\label{tab:images}
\end{table}

\subsection{VAW-CZSL}
The VAW dataset \cite{vaw-czsl} is a multi-label attribute and object dataset that serves as the source for VAW-CZSL. It contains a total of 541 objects and 440 states. The training set comprises over 72,000 images and includes 11,175 compositions. For evaluation, the validation set has about 9,500 images covering 2,322 unseen compositions, while the test set contains approximately 11,000 images featuring 2,470 unseen compositions. VAW-CZSL represents the largest available dataset, characterized by its extensive collection of attributes and objects. Even though it is a recent dataset, its considerable scale often requires substantial computational power, particularly for the implementation of hybrid approaches, thereby restricting its widespread application in current research. Detailed statistics are presented in Table \ref{tab:dataset}.


\subsection{Clothing16K}
Clothing16K is a dataset focused on clothing, featuring clothing objects (e.g., ``Shirt'') and color attributes (e.g., ``black''). It includes 9 attributes and 8 objects. The dataset is split into training, validation, and test sets. The training set comprises over 7,000 images and 18 compositions. The validation set contains over 5,500 images with 10 unseen compositions, while the test set has around 3,500 images featuring 8 unseen compositions. Clothing16K is a simpler and smaller dataset. While it was utilized by early works, it is no longer commonly employed in recent research. The details of the Clothing16K dataset are presented in Table \ref{tab:dataset}.

\subsection{MAD}

The Multi-Attribute Dataset (MAD) \cite{SMAO} is a dataset featuring multiple attributes. It consists of 158 attributes and 309 objects. The training set comprises over 81,000 images and includes 5,630 compositions. For evaluation, the validation set contains over 27,000 images with 1000 unseen compositions, while the test set has 42,000 images featuring 1,400 unseen compositions. MAD is the only dataset structured with multiple attributes for a single object. This makes it an ideal choice for anyone looking to work with multi-attribute problems, rather than the more traditional single-attribute and object setups. Details about the MAD dataset are presented in Table \ref{tab:dataset}.

\subsection{AO-CLEVr}
AO-CLEVr is a synthetic dataset designed for Compositional Zero-Shot Learning. It features 3 types of objects (sphere, cube, cylinder) and 8 attributes (red, purple, yellow, blue, green, cyan, gray, and brown). Notably, this is the smallest dataset available for CZSL research. \\

In summary, UT-Zappos, MIT-States, and CGQA are the most widely used datasets in CZSL. UT-Zappos provides a clean and well-structured benchmark for fast testing, but its relatively small set of attributes and objects restricts variability, making it less indicative of a model’s generalization ability. By contrast, MIT-States and CGQA offer greater diversity and contextual complexity, providing a stronger basis for testing compositional reasoning. Some recent works have also reported results on WAV-CZSL, though its use remains less common. Building on these datasets, different evaluation protocols have been developed to assess models under both closed-world and open-world settings, which we review in the next section.

\section{Evaluation Protocols}
\label{sec:evaluation}

\subsection{Area Under the Curve}
Evaluating compositional zero-shot learning methods using only simple accuracy on seen or unseen compositions is often insufficient due to the inherent bias favoring training compositions. To provide a more comprehensive assessment, Area Under the Curve (AUC) serves as an important metric. It measures the capacity of the model to effectively distinguish between seen and unseen compositions across a range of bias thresholds. Bias is defined as a scalar value added to the prediction score of novel (unseen) compositions during the inference phase. By systematically adjusting the bias value, one can observe the performance trade-off between the accuracy of the model on seen and unseen compositions. This relationship is typically visualized by plotting unseen accuracy against seen accuracy across these different bias values. The AUC represents the area beneath this curve, providing a single aggregate value that encapsulates the performance of the model across all bias levels and thus offering a robust indicator of its generalized recognition ability.

\subsection{Seen Accuracy}
Seen accuracy evaluates the performance of the model specifically on the subset of the test data comprising seen compositions. In Generalized closed-world CZSL (GCZSL), where models must classify both training (seen) and novel (unseen) compositions, assessing seen accuracy is crucial. It ensures that the model effectively retains its ability to recognize compositions it was trained on, preventing catastrophic forgetting or undue bias towards unseen categories. This metric, alongside unseen accuracy, provides critical insight into the model's balance between generalization to new concepts and performance on established knowledge, serving as a vital component in calculating the overall Harmonic Mean.

The highest seen accuracy is generally attained when the bias term, applied to the prediction scores of novel (unseen) compositions during inference, is set to a sufficiently negative value. This negative bias serves to depress the scores of all unseen candidates relative to seen ones, effectively biasing the predictions of the model towards the seen composition space. This strategy maximizes the likelihood of correctly classifying images belonging to seen compositions, though it often results in diminished performance on unseen compositions.

\subsection{Unseen Accuracy}
Unseen accuracy quantifies the performance of the model specifically on the subset of the test data comprising unseen compositions. This metric is fundamental to the evaluation of CZSL, as it directly measures the generalization capability of the model, which is its core objective. It provides crucial insight into an ability of a model to predict novel compositions that it has never encountered during training. In the context of generalized closed-world, unseen accuracy, when considered alongside seen accuracy, helps to reveal the inherent trade-offs of the model and is a necessary component for calculating the overall Harmonic Mean.

Optimal unseen accuracy is generally attained when the bias term, applied to the prediction scores of novel (unseen) compositions, is maximized (assigned a large positive value). The application of a substantial positive bias results in a significant increase in the scores of all unseen composition candidates relative to their seen counterparts. This strong positive bias effectively directs the model to favor the prediction of unseen labels, thereby enhancing the probability of correctly identifying the ground truth label for images belonging to the unseen composition set. Under this configuration, where the model is heavily biased towards unseen predictions, the evaluation scenario becomes functionally equivalent to that of standard zero-shot learning, which exclusively evaluates performance on novel classes.

\subsection{Harmonic Mean}

The Harmonic Mean is a metric used to evaluate the overall performance of a model by capturing the balance between its seen and unseen accuracy (Acc). It is particularly useful and appropriate in Compositional Zero-Shot Learning, especially in the generalized closed-world setting, because models must perform effectively on both training (seen) and novel (unseen) compositions. The Harmonic Mean intrinsically penalizes models that exhibit a significant disparity in performance between these two splits, thus strongly favoring models that achieve a robust and equitable trade-off. This characteristic makes it a more reliable indicator of true generalization capability of the model compared to simply averaging individual accuracies, as it reflects a holistic ability to handle the full spectrum of compositional challenges.
The Harmonic Mean is defined by the formula: 
$$
\text{Harmonic Mean} = \frac{(2)  (\text{Acc}_\text{seen})  (\text{Acc}_\text{unseen})}{\text{Acc}_\text{seen} + \text{Acc}_\text{unseen}}.
$$ 
The best Harmonic Mean is achieved when the seen and unseen accuracies are high and close to each other, representing a successful trade-off.

{\small
\begin{table}
  \caption{Table presents the performance of various state-of-the-art models across three common CZSL datasets. Results for the closed-world setting are presented in the upper section of the table while the open-world setting results are in the lower section. Methods are further organized by their visual backbone within each setting, showing results for models utilizing a ResNet backbone followed by those using the Visual Transformer (ViT) and Clip encoder. Performance is evaluated using Seen accuracy (S), Unseen accuracy (U), and Harmonic Mean (HM). An asterisk (*) indicates that the backbone was fine-tuned during training. All results are reproduced from the respective original papers. Top performance in each column is highlighted in \textbf{bold}, while the second-best performance is shown in \textcolor{blue}{blue.}}
  \label{tab:results}
  \setlength{\tabcolsep}{3.3pt}
  \begin{tabular}{l|c|cccc|cccc|cccc}
    \toprule
    \multirow{2}{*}{\textbf{Models}} & \multirow{2}{*}{\textbf{Backbone}} 
    & \multicolumn{4}{c|}{\textbf{MIT-States}} 
    & \multicolumn{4}{c|}{\textbf{UT-Zappos}} 
    & \multicolumn{4}{c}{\textbf{CGQA}} \\
    & & AUC & HM & S & U & AUC & HM & S & U & AUC & HM & S & U \\
    \hline
     \multicolumn{14}{c}{\textbf{Closed-world}} \\
    \hline
    AoP \cite{nagarajan2018attributes} & ResNet & 1.6 &  9.9 & 14.3 & 17.4 & 25.9 & 40.8 & 59.8 & 54.2 & 0.7 & 5.9 & 17.0 & 5.6 \\ 
    Causal \cite{atzmon2020causal} & ResNet & - & - & - & - &  23.3 & 31.8 & 39.7 & 26.6 & - & - & - & - \\
    
    SCEN \cite{SCEN}& ResNet & 5.3 & 18.4 & 29.9 & 25.2 & 32.0 & 47.8 & 63.5 & 63.1 & 5.5 & 17.5 & 28.9 & 25.4 \\

    Compcos \cite{mancini2021open} & ResNet &  4.5 & 16.4 & 25.3 & 24.6 & 28.7 & 43.1 & 59.8 & 62.5 & 2.6 & 12.4 & 28.1 & 11.2 \\
    CGE* \cite{naeem2021learning} & ResNet & 6.5 & 21.4 & 32.8 & 28.0 & 33.5 & \textbf{60.5} & 64.5 & 71.5 & 4.2 & 16.0 & 33.5 & 15.5 \\
    Co-CGE* \cite{mancini2022learning} & ResNet & 6.6 & 20.0 & 32.1 & 28.3 & 33.9 & 48.1 & 62.3 & 66.3 & 4.1 & 15.5 & 33.3 & 14.9 \\
    DeCa* \cite{DeCa} & ResNet & 6.6 & 20.3 & 32.2 & 27.4 & 37.0 & 51.7 & 64.0 & 68.8 & - & - & - & - \\
    OADis \cite{OADis} & ResNet & 5.9 & 18.9 & 31.1 & 25.6 & 30.0 & 44.4 & 59.5 & 65.5 & - & - & - & -  \\
    CoT* \cite{CoT} & ViT & 10.5 & 25.8 & 39.5 & 33.0 & - & - & - & - &  7.4 & 22.1 & 39.2 & 22.7 \\
    CSP \cite{csp2023} & Clip & 19.4 & 36.3 & 46.6 & 49.9 & 33.0 & 46.6 & 64.2 & 66.2 & 6.2 & 20.5 & 28.8 & 26.8 \\
    GIPCOL \cite{Gipcol} & Clip & 19.9 & 36.6 & 48.5 & 49.6 & 36.2 & 48.8 & 65.0 & 68.5 & 7.1 & 22.5 & 31.9 & 28.4 \\
    Dynamic \cite{hu2024dynamic} & Clip & 20.0 & 37.4 & 46.3 & 49.8 & 39.6 & 52.0 & 67.1 & 72.5 & 7.3 & 21.9 & 32.4 & 28.5 \\
    HPL \cite{HPL} & Clip & 20.2 & 37.3 & 47.5 & 50.6 & 35.0 & 48.2 & 63.0 & 68.8 & 7.2 & 22.4 & 30.8 & 28.4 \\

    DFSP \cite{DFSP} & Clip & 20.6 & 37.3 & 46.9 & 52.0 & 36.0 & 47.2 & 66.7 & 71.7 & 10.5 & 27.1 & 38.2 & 32.0  \\
    
    Troika \cite{huang2024troika} & Clip & 22.1 & 39.3 & 49.0 & 53.0 & 41.7 & 54.6 & 66.8 & 73.8 & 12.4 & 29.4 & 41.0 & 35.7  \\
    PLID \cite{PLID} & Clip & 22.1 & 39.0 & 49.7 & 52.4 & 38.7 & 52.4 & 49.7 & 52.4 & 11.0 & 27.9 & 38.8 & 33.0 \\
    CDS-CZSL \cite{CDS-CZSL} & Clip & 22.4 & 39.2 & 50.3 & 52.9 & 39.5 & 52.7 & 63.9 & \textcolor{blue}{74.8} & 11.1 & 28.1 & 38.3 & 34.2  \\
    
    Retrieval \cite{Retrieval} & Clip & 22.5 & 39.2 & 50.0 & 53.3 & \textcolor{blue}{44.5} & 56.5 & \textcolor{blue}{69.4} & 72.8 & 14.4 & 32.0 & \textbf{45.6} & 36.0 \\

    CAILA \cite{caila} & Clip & \textcolor{blue}{23.4} & \textcolor{blue}{39.9} & \textcolor{blue}{51.0} & \textcolor{blue}{53.9} & 44.1 & 57.0 & 67.8 & 74.0 & \textcolor{blue}{14.8} & \textcolor{blue}{32.7} & 43.9 & \textbf{38.5} \\

    CLUSPRO \cite{CLUSPRO} & Clip & \textbf{23.8} & \textbf{40.7} & \textbf{52.1} & \textbf{54.0} & \textbf{46.6} & \textcolor{blue}{58.5} & \textbf{70.7} & \textbf{76.0} & \textbf{14.9} & \textbf{32.8} & \textcolor{blue}{44.3} & \textcolor{blue}{37.8} \\

    \hline
    \multicolumn{14}{c}{\textbf{Open-world}} \\
    \hline
    KG-SP* \cite{kg-sp} & ResNet & 1.3 & 7.4 & 28.4 & 7.5 & 26.5 & 42.3 & 61.8 & 52.1 & 0.78 & 4.7 & 31.5 & 2.9 \\
    ASP* \cite{asp} & ResNet & 1.4 & 7.7 & 27.1 & 8.4 & 25.9 & 43.1 & 61.0 & 48.6 & 0.80 & 5.0 & 31.7 & 3.2 \\
    CompCos \cite{mancini2021open} & ResNet & 1.6 & 8.9 & 25.4 & 10.0 & 21.3 & 36.9 & 59.3 & 46.8 & 0.39 & 2.8 & 28.4 & 1.8 \\
    Co-CGE* \cite{mancini2022learning} & ResNet & 2.3 & 10.7 & 30.3 & 11.2 & 23.3 & 40.8 & 61.2 & 45.8 & 0.78 & 4.8 & 32.1 & 3.0 \\
    CSP \cite{csp2023} & Clip & 5.7 & 17.4 & 46.3 & 15.7 & 22.7 & 38.9 & 64.1 & 44.1 & 1.20 & 6.9 & 28.7 & 5.2 \\
    GIPCOL \cite{Gipcol} & Clip & 6.3 & 17.9 & 48.5 & 16.0 & 23.5 & 40.1 & 65.0 & 45.0 & 1.30 & 7.3 &  31.6 & 5.5  \\
    
    HPL \cite{HPL} & Clip & 6.9 & 19.8 & 46.4  & 18.9 & 24.6  & 40.2 & 63.4 & 48.1 & 1.37 & 7.5 & 30.1  & 5.8 \\

    DFSP \cite{DFSP} & Clip & 6.8 & 19.3 & 47.5 & 18.5 & 30.3 & 44.0 & 66.8 & 60.0 & 2.40 & 10.4 & 38.3 & 7.2  \\
    
    Troika \cite{huang2024troika} & Clip & 7.2 & 20.1 & 48.8 & 18.7 & 33.0 & 47.8 & 66.4 & 61.2 & 2.70 & 10.9 & 40.8 & 7.9  \\
    PLID \cite{PLID} & Clip & 7.3 & 20.4 & 49.1 & 18.7 & 30.8 & 46.6 & 67.6 & 55.5 & 2.50 & 10.6 & 39.1 & 7.5 \\
    CDS-CZSL \cite{CDS-CZSL} & Clip & \textcolor{blue}{8.5} & \textcolor{blue}{22.1} & 49.4 & \textcolor{blue}{21.8} & 32.3 & 48.2 & 64.7 & \textcolor{blue}{61.3} & 2.68 & \textcolor{blue}{11.6} & 37.6 & 8.2  \\
    
    Retrieval \cite{Retrieval} & Clip & 8.2 & 21.8 & 49.9 & 20.1 & \textcolor{blue}{33.3} & 47.9 & \textcolor{blue}{69.4} & 59.4 & \textbf{4.4} & \textbf{14.6} & \textbf{45.5} & \textbf{11.2} \\
    
     CAILA \cite{caila} & Clip & 8.2 & 21.6 & \textcolor{blue}{51.0} & 20.2 & 32.8 & \textcolor{blue}{49.4} & 67.8 & 59.7 & \textcolor{blue}{3.08} & 11.5 & \textcolor{blue}{43.9} & 8.0 \\ 

     CLUSPRO \cite{CLUSPRO} & Clip & \textbf{9.3} & \textbf{23.0} & \textbf{51.2} & \textbf{22.1} & \textbf{39.5} & \textbf{54.1} & \textbf{71.0} & \textbf{66.2} & 3.0 & \textcolor{blue}{11.6} & 41.6 & \textcolor{blue}{8.3} \\
  \bottomrule
\end{tabular}
\end{table}
}

\section{Empirical Evaluation and Comparative Analysis}
\label{sec:performance}


To complement the taxonomy of CZSL methods presented in the section \ref{sec:taxonomy}, we now examine their empirical performance. Organizing results by taxonomy allows us to assess how different disentanglement strategies translate into measurable gains in compositional generalization. Performance is reported under two standard evaluation protocols: the closed-world setting, where the set of possible attribute-object pairs is fixed, and the open-world setting, where the label space expands to all combinations of training attributes and objects, including unfeasible pairs. This dual perspective provides insight into both controlled compositional reasoning and more realistic, challenging scenarios. The following subsections present results for each setting, highlighting trends that mirror the conceptual progression outlined in our taxonomy.

\subsection{\textbf{Closed-world Setting}}
In the closed-world setting, models are evaluated on a fixed set of test compositions, including both seen and unseen pairs. A first and most striking trend across all categories is the backbone effect: earlier methods using ResNet encoders consistently achieve lower performance, while more recent approaches built on CLIP encoders (from 2023 onwards) show a substantial improvement in accuracy. This shift reflects the field’s transition toward vision-language pretraining as the standard backbone, providing stronger multimodal grounding before disentanglement strategies are even applied. In light of these patterns, we now analyze the results category by category according to our taxonomy.

\subsubsection*{No Explicit Disentanglement}

    The earlier CZSL approaches fall under this category, and their results confirm the limitations highlighted in our taxonomy. By treating attribute-object pairs holistically without explicit primitive separation, these models consistently occupy the lower end of the performance spectrum as shown in Figures (\ref{fig:scatter-mit-closed}, \ref{fig:scatter-utzappos-closed}, \ref{fig:scatter-cgqa-closed}). Their modest AUC scores illustrate the weaknesses of composition-as-class strategies, which lack scalability and generalization power when tested with novel attribute-object combinations. Within ResNet-based methods, \textbf{context-aware models} \textit{such as CGE \cite{naeem2021learning} and Co-CGE \cite{mancini2022learning} achieve stronger results} than their counterparts by leveraging graph neural networks to capture richer relational structures. While ResNet-based models remain limited, the \textit{Dynamic \cite{hu2024dynamic} model, built on CLIP and trained across multiple domains, achieves the highest accuracy in this category} by leveraging richer contextual diversity for better generalization.

\subsubsection*{Textual Disentanglement}
\begin{figure*}[htbp] 
    \centering
    \subfloat[AUC of representative methods on MIT-States dataset in the closed-world setting.]{\includegraphics[width=0.49\textwidth]{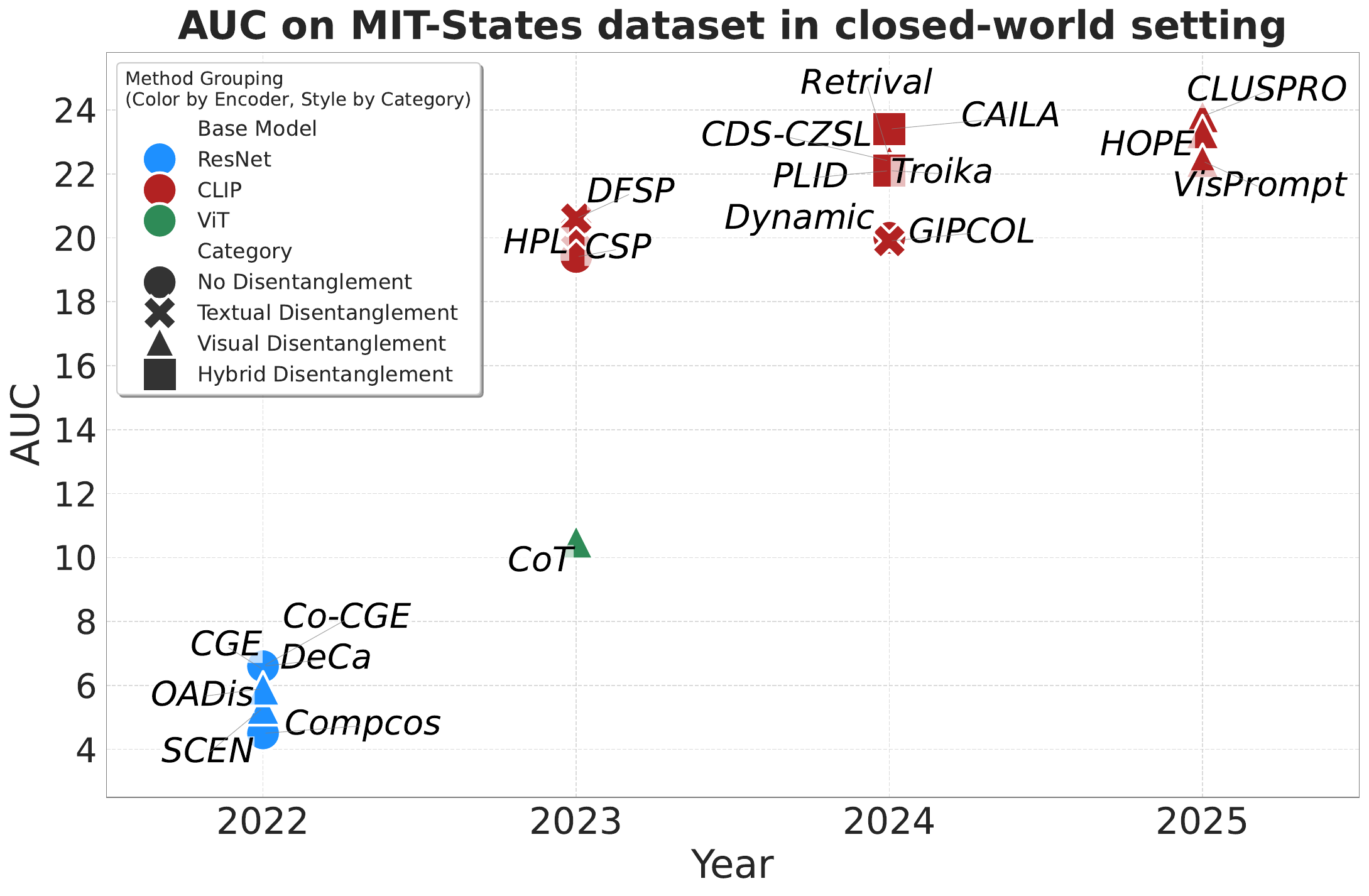}\label{fig:scatter-mit-closed}} \hfill
    \subfloat[AUC of representative methods on MIT-States dataset in the open-world setting.]{\includegraphics[width=0.49\textwidth]{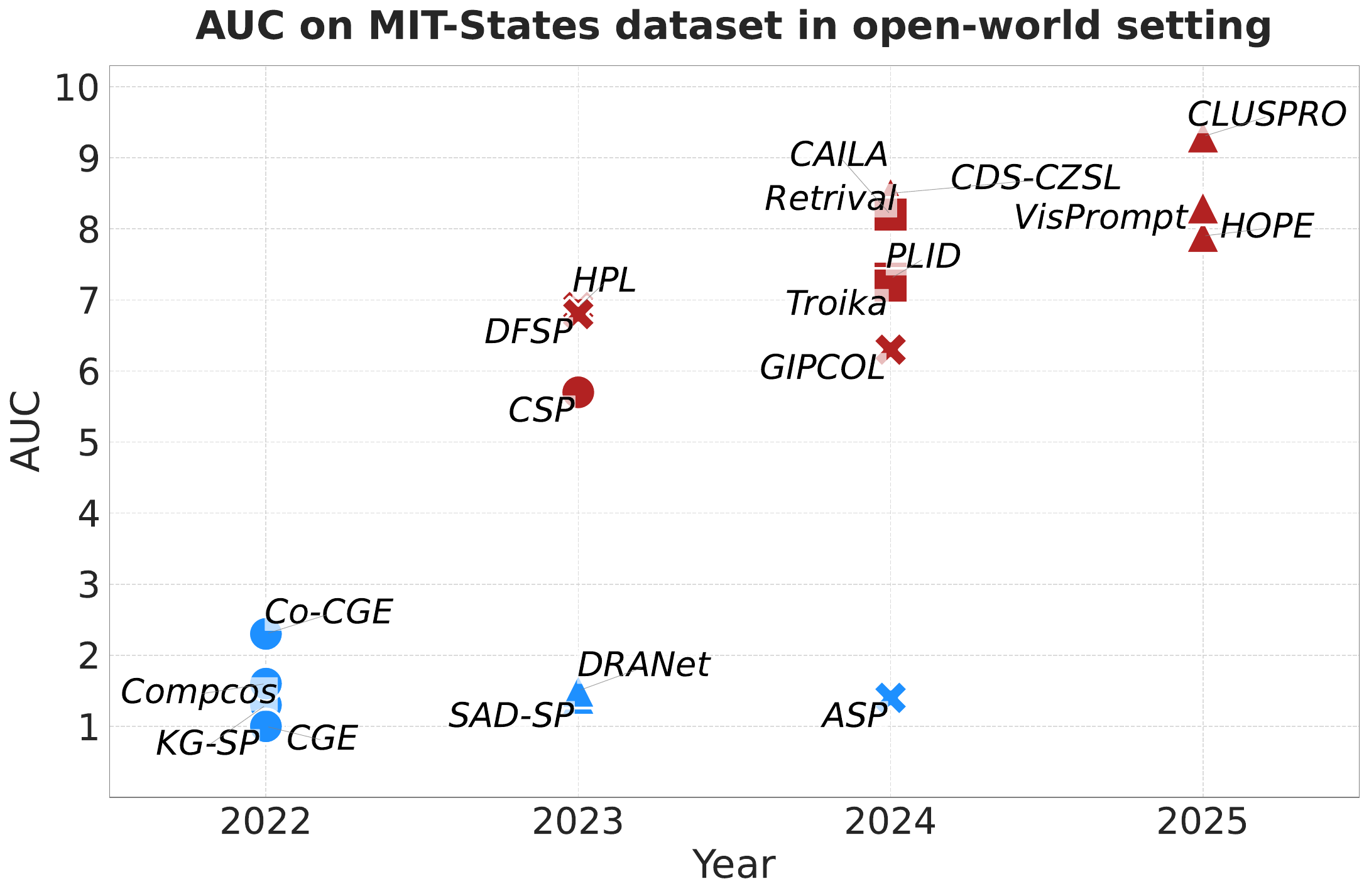}\label{fig:scatter-mit-open}} \\
    \subfloat[AUC of representative methods on UT-Zappos dataset in the closed-world setting.]{\includegraphics[width=0.49\textwidth]{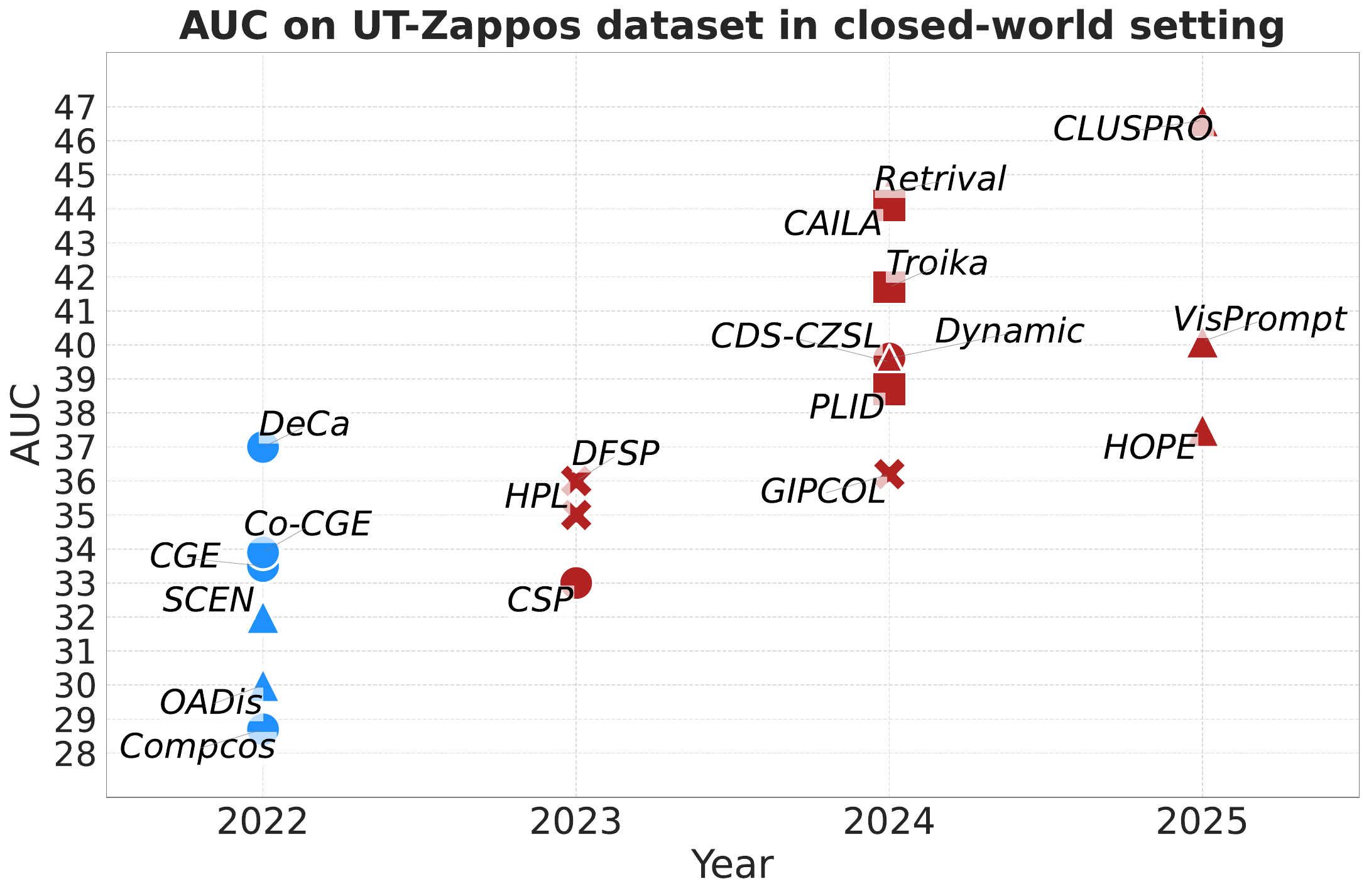}\label{fig:scatter-utzappos-closed}} \hfill
    \subfloat[AUC of representative methods on UT-Zappos dataset in the open-world setting.]{\includegraphics[width=0.49\textwidth]{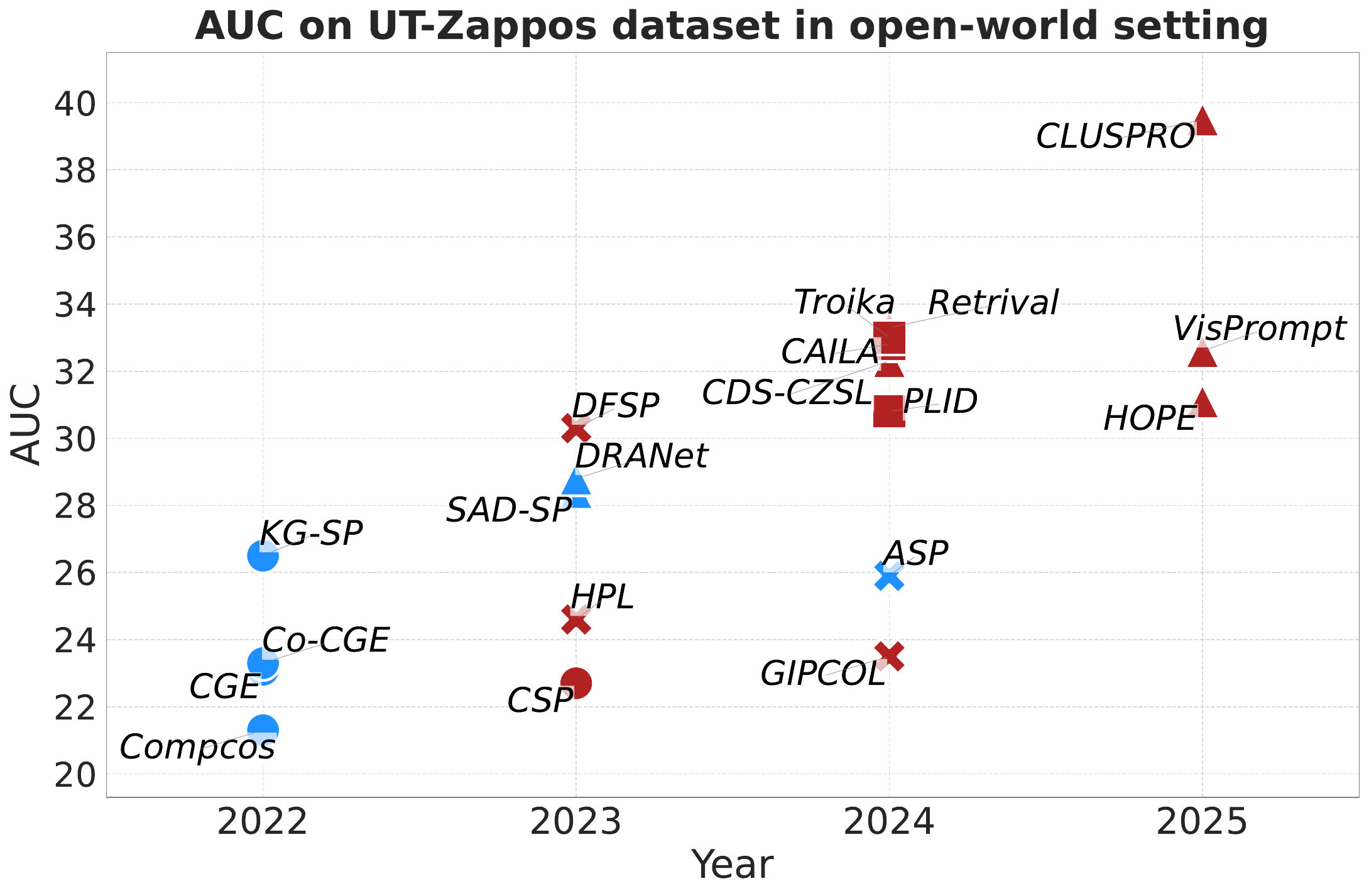}\label{fig:scatter-ut-zappos-open}} \\
    \subfloat[AUC of representative methods on CGQA dataset in the closed-world setting.]{\includegraphics[width=0.49\textwidth]{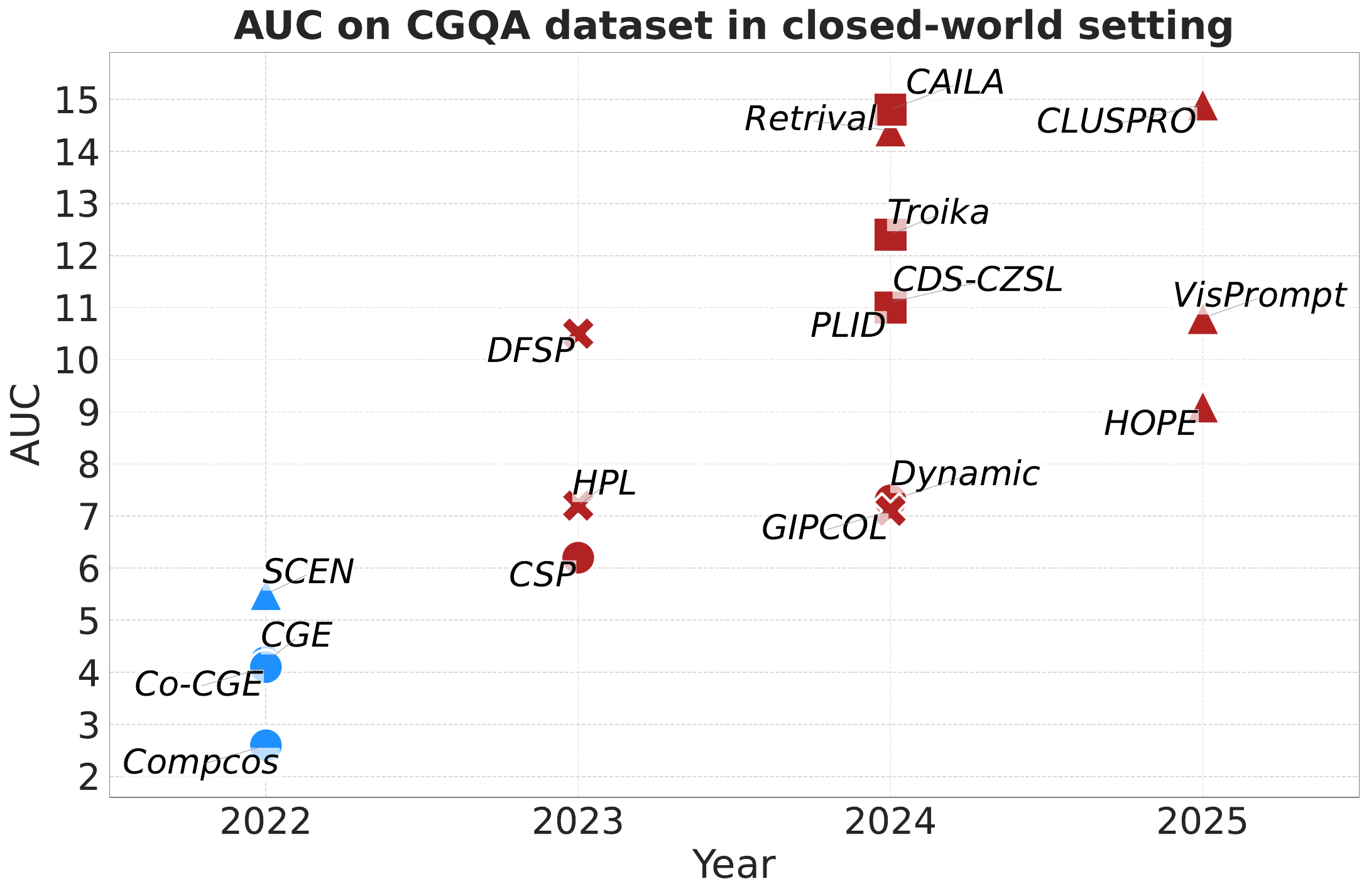}\label{fig:scatter-cgqa-closed}} \hfill
    \subfloat[AUC of representative methods on CGQA dataset in the open-world setting.]{\includegraphics[width=0.49\textwidth]{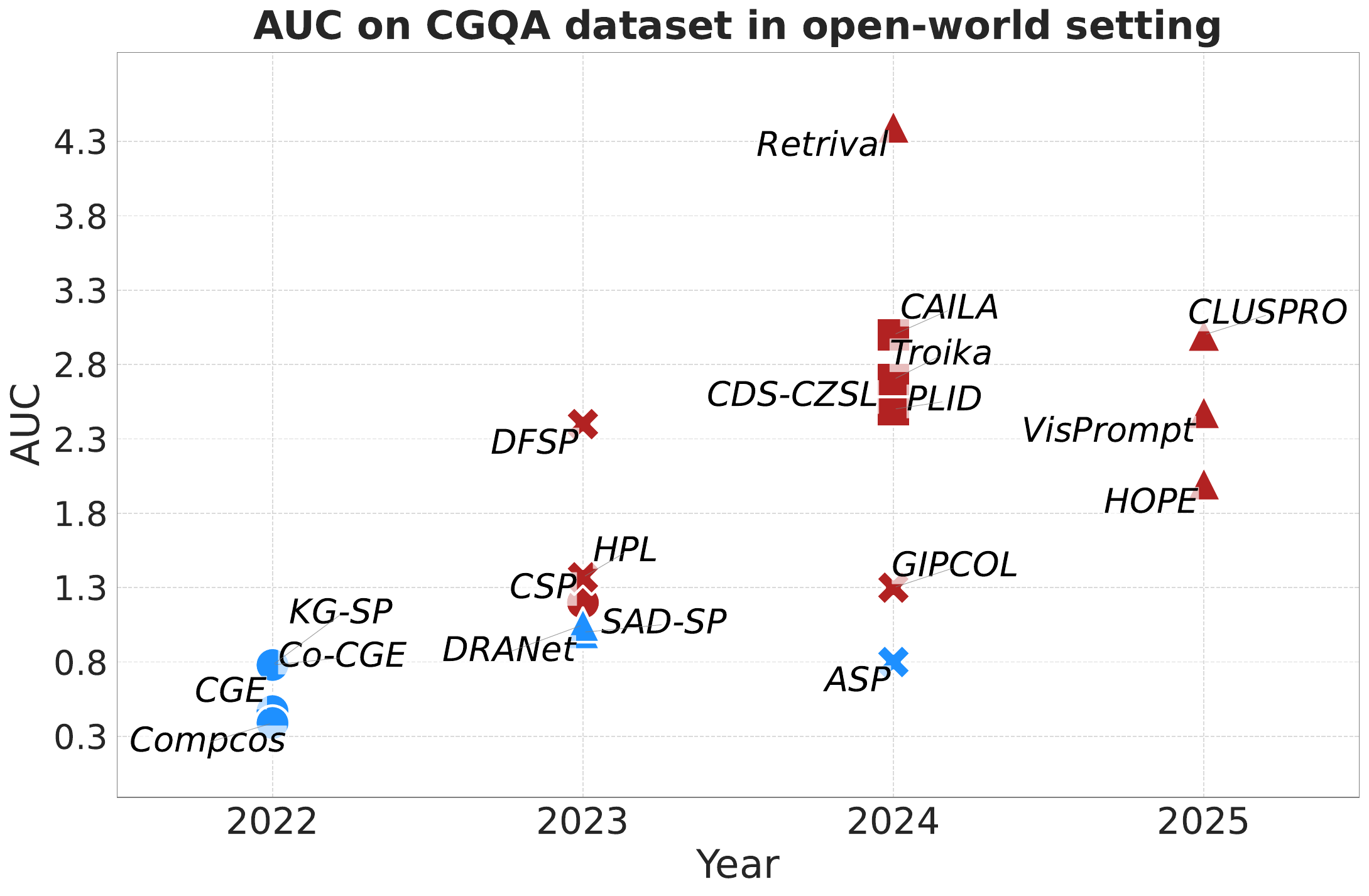}\label{fig:scatter-cgqa-open}}
    \caption{Graphs showing the AUC scores of CZSL methods on the MIT-States, UT-Zappos and CGQA datasets in closed-world and open-world settings. Each data point in the graph is uniquely defined by both its \textbf{visual encoder (indicated by color)} and its \textbf{disentanglement category (indicated by shape)}. ResNet-based models are shown in \textcolor{blue}{blue}, ViT-based models are shown in \textcolor{green}{green} and Clip-based models in \textcolor{red}{red}. The shape of each point indicates the disentaglement category: a circle \includegraphics[width=0.1in, height=0.1in]{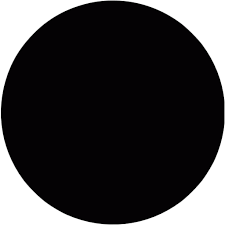} for No Disentanglement, a cross Textual Disentanglement by \includegraphics[width=0.1in, height=0.1in]{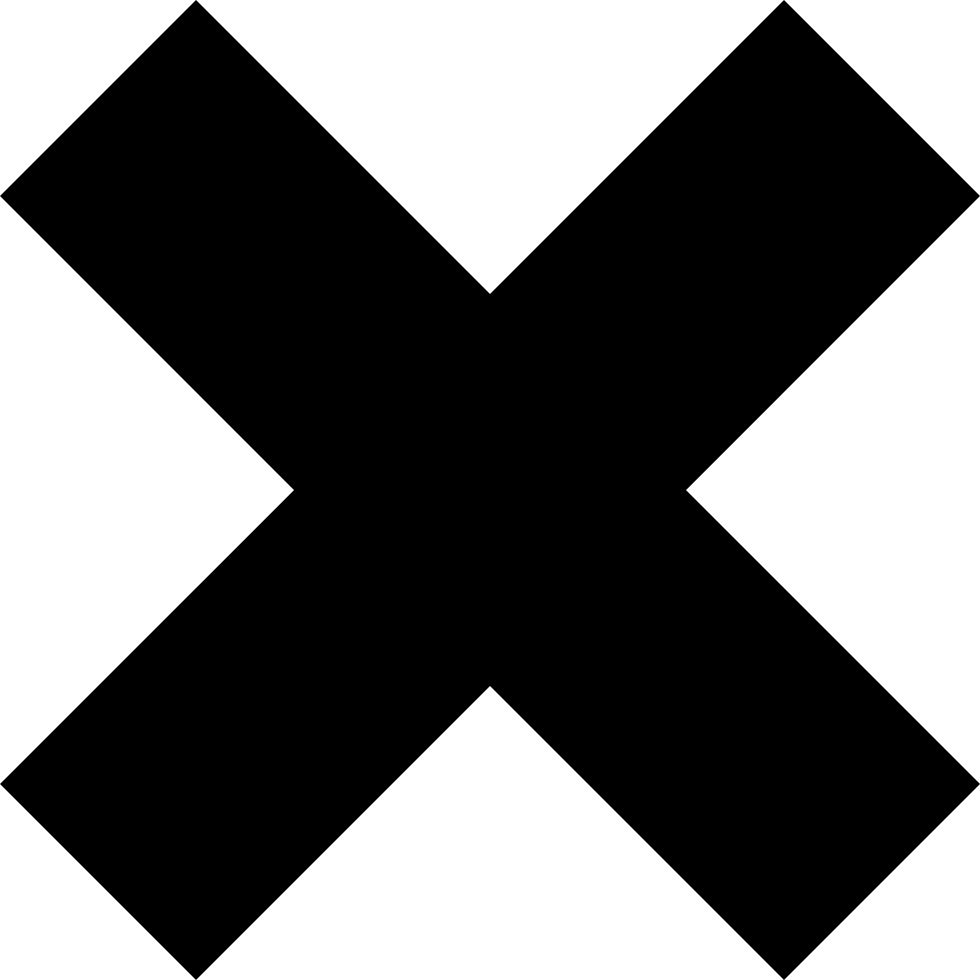} for Textual Disentanglement, a triangle \includegraphics[width=0.1in, height=0.1in]{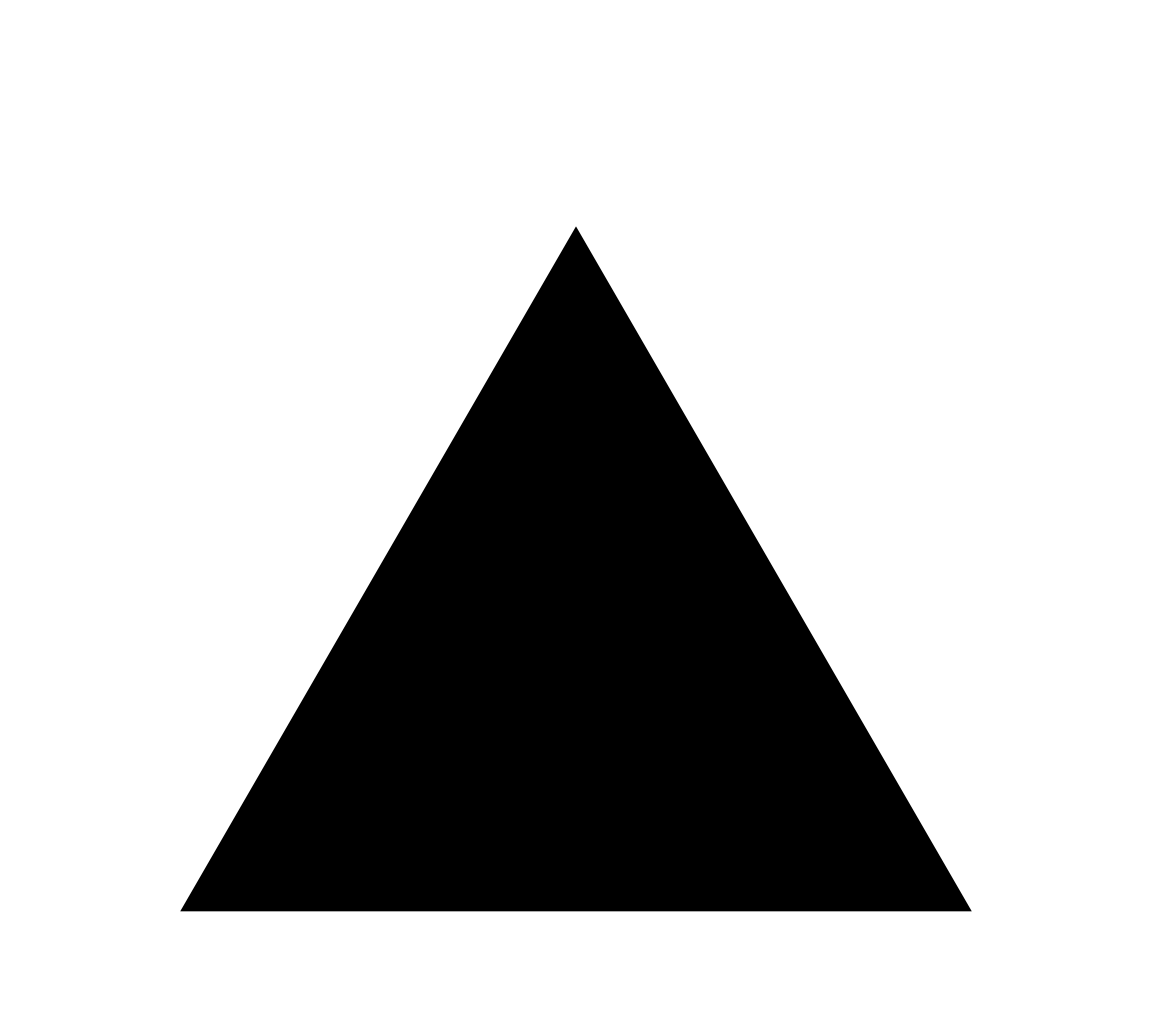} for Visual Disentanglement, and a square \includegraphics[width=0.1in, height=0.1in]{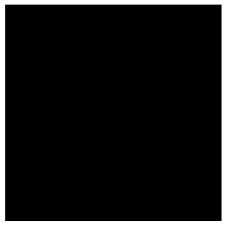} for Cross-Model (Hybrid) Disentanglement.}
    \label{fig:all_plots}
\end{figure*}

Relatively fewer methods have pursued textual disentanglement, yet they achieve noticeable improvements over no-disentanglement baselines as shown in Figures (\ref{fig:scatter-mit-closed}, \ref{fig:scatter-utzappos-closed}, \ref{fig:scatter-cgqa-closed}). This trend validates the intuition that attributes and objects, already separable in linguistic space, can provide a stronger starting point for compositional reasoning. Within this category, however, performance differs across subcategories. \textbf{Primitive-aware methods}, such as DFSP \cite{DFSP}, \textit{achieve stronger results} by incorporating individual attribute and object embeddings directly into the composition score, ensuring that primitive-level knowledge contribute explicitly to composition prediction. In contrast, pairwise-composition methods, such as GIPCOL \cite{Gipcol}, rely solely on holistic attribute-object embeddings, which limits their ability to transfer knowledge across pairs. Despite these gains, textual disentanglement as a whole still lags behind visual disentanglement methods, underscoring the insufficiency of language alone to capture the appearance-level variability that characterizes visual primitives.

\subsubsection*{Visual Disentanglement}
The majority of recent methods belong to this category, reflecting the field’s recognition that visual entanglement poses the most pressing challenge for CZSL. As predicted by the taxonomy and shown in Figures (\ref{fig:scatter-mit-closed}, \ref{fig:scatter-utzappos-closed}, \ref{fig:scatter-cgqa-closed}), explicitly separating attributes and objects in visual space yields clear gains over textual approaches. Within this broad category, certain subcategories stand out. \textbf{Augmentation-enhanced methods} such as Retrieval \cite{Retrieval} benefit from enriching primitive variability during training, while \textbf{prototype-centric methods} like CLUSPRO \cite{CLUSPRO} leverage clustering of attributes and objects to establish stable anchors for composition. Similarly, \textbf{conditional attribute modeling} approaches such as CDS-CZSL \cite{CDS-CZSL} adapt attributes to their object contexts, capturing fine-grained variability more effectively. Among these, \textit{CLUSPRO achieves the highest accuracy across all three benchmark datasets}, outperforming not only other visual disentanglement models but also cross-modal approaches. These results highlight both the diversity and maturity of visual disentanglement methods, though the gains remain uneven across subcategories.

\subsubsection*{Cross-Modal Disentanglement}
Hybrid approaches that align visual and textual primitives have only recently emerged, with most contributions appearing in 2024. Despite their small number, \textit{these methods already demonstrate strong results, often on par with or exceeding the best visual models} in datasets such as MIT-States and CGQA. Within this group, \textbf{LLM-guided prompting} methods such as \textit{PLID \cite{PLID} show potential but achieve lower accuracy than \textbf{encoder-augmented approaches} like CAILA \cite{caila} and Troika \cite{huang2024troika}, which refine primitive representations more effectively within the backbone.} Interestingly, CLUSPRO, a prototype-centric visual method that clusters attributes and objects to build more stable primitive anchors, achieves the highest accuracy on UT-Zappos. However, on more diverse datasets such as MIT-States and CGQA, CAILA matches its performance, underscoring the advantage of cross-modal alignment when greater contextual variability is present. Overall, this line of work is still in its early stages but evolving rapidly, with clear evidence of great potential. More contributions are needed to explore its true capacity and establish cross-modal disentanglement as the dominant paradigm for robust compositional generalization. \\

In the closed-world setting, the results across categories closely follow the progression outlined in our taxonomy. The earliest \textbf{no-disentanglement} methods remain at the lower end of the spectrum, highlighting the limitations of treating compositions holistically. \textbf{Textual disentanglement} improves upon these baselines but is constrained by its reliance on language alone, which cannot capture the variability of visual attributes. \textbf{Visual disentanglement} emerges as the most developed line of work, with approaches such as CLUSPRO and Retrieval achieving state-of-the-art performance and, in some cases, rivaling hybrid approaches. Finally, \textbf{cross-modal disentanglement}, although still limited in number, shows the strongest potential: encoder-augmented methods like CAILA and Troika already match visual baselines on diverse datasets, signaling that multimodal alignment is likely to drive the next wave of progress. Overall, the closed-world results from Table \ref{tab:results} and Figures (\ref{fig:scatter-mit-closed}, \ref{fig:scatter-utzappos-closed}, \ref{fig:scatter-cgqa-closed}) illustrate a clear trajectory from holistic to unimodal disentanglement and now toward cross-modal integration as the most promising direction.

\subsection{\textbf{Open-world Setting}}
In the open-world setting, models are evaluated over all combinations of training attributes and objects, including unfeasible pairs. This expanded label space makes the task significantly more challenging than the closed-world setting, as models must simultaneously handle recognition and feasibility. Similar to the closed-world results, CLIP-based methods consistently outperform ResNet-based approaches, confirming the advantage of vision-language pretraining even under the more demanding open-world evaluation.

\subsubsection*{No Explicit Disentanglement}
In the open-world setting, no-disentanglement methods remain the weakest group overall, yet some interesting trends emerge. Similar to the closed-world case, \textbf{context-aware models} \textit{such as Co-CGE \cite{mancini2022learning} perform better than other ResNet-based approaches} by leveraging relational structure as shown in Figures (\ref{fig:scatter-mit-open}, \ref{fig:scatter-ut-zappos-open}, \ref{fig:scatter-cgqa-open}). However, a different pattern appears in this category under open-world setting. KG-SP \cite{kg-sp}, which predicts primitives independently rather than modeling compositions holistically, performs on par with Co-CGE on CGQA, surpasses it on UT-Zappos, but falls behind on MIT-States. \textit{This variation suggests that in the open-world setting, where the test label space expands to include all possible attribute-object combinations, even simple primitive-level predictions can sometimes rival or exceed more sophisticated context-aware strategies}, depending on dataset characteristics.

\subsubsection*{Textual Disentanglement}
In the open-world setting, textual disentanglement methods continue to outperform no-disentanglement baselines, but their relative strengths vary across subcategories. As illustrated in Figures (\ref{fig:scatter-mit-open}, \ref{fig:scatter-ut-zappos-open}, \ref{fig:scatter-cgqa-open}), \textbf{primitive-aware methods} such as DFSP \cite{DFSP} achieve consistently higher accuracy than pairwise-composition methods like GIPCOL \cite{Gipcol}. Notably, the performance gap between these two approaches is larger than in the closed-world setting, indicating that \textit{explicitly incorporating primitive-level information into the prediction score becomes even more beneficial when models must generalize across the entire space of attribute-object combinations.}

\subsubsection*{Visual Disentanglement}
In the open-world setting, visual disentanglement remains the most active and competitive category, showing clear gains over textual methods. As seen in Figures (\ref{fig:scatter-mit-open}, \ref{fig:scatter-ut-zappos-open}, \ref{fig:scatter-cgqa-open}), \textbf{augmentation-enhanced methods} such as Retrieval \cite{Retrieval}, \textbf{prototype-centric approaches} like CLUSPRO \cite{CLUSPRO}, and \textbf{conditional attribute models} such as CDS-CZSL \cite{CDS-CZSL} continue to perform at the top. Interestingly, \textbf{auxiliary-objective methods} such as VisPrompt \cite{VisPrompt} also achieve results on par with these leading subcategories, demonstrating that \textit{auxiliary constraints can be as effective as architectural disentanglement strategies under the expanded label space in the open-world setting}. Dataset-specific trends are also evident: on CGQA, the largest benchmark, \textit{Retrieval \cite{Retrieval} achieves the highest accuracy by a good margin}; on MIT-States, CLUSPRO \cite{CLUSPRO} leads the group; and on UT-Zappos, CLUSPRO again attains the best performance. These results highlight both the maturity and diversity of visual disentanglement methods, as well as their robustness across different dataset scales and attribute distributions.

\subsubsection*{Cross-Modal Disentanglement}
While cross-modal approaches matched or even surpassed the best visual disentanglement methods in the closed-world setting, a different trend emerges in the open-world evaluation. As shown in Figures (\ref{fig:scatter-mit-open}, \ref{fig:scatter-ut-zappos-open}, \ref{fig:scatter-cgqa-open}), there is a noticeable performance gap, with a few visual disentanglement methods consistently outperforming cross-modal approaches across datasets. \textit{This suggests that under the expanded label space, visual grounding alone remains more reliable than joint visual-textual alignment, which may be more susceptible to noise introduced by infeasible combinations.} Among the cross-modal models, CAILA \cite{caila} achieves the strongest results, outperforming alternatives such as PLID \cite{PLID} and Troika \cite{huang2024troika}, yet still falling short of top-performing visual methods like CLUSPRO and Retrieval. Overall, these findings indicate that while cross-modal disentanglement holds great promise, more research is needed to refine these methods and fully realize their potential in open-world CZSL. \\

Overall, in the open-world setting, performance trends diverge from those observed under closed-world evaluation. \textbf{No-disentanglement methods} remain the weakest group overall, though primitive-level strategies such as KG-SP occasionally rival context-aware models depending on the dataset. \textbf{Textual disentanglement} continues to provide gains over these baselines, with primitive-aware approaches like DFSP outperforming pairwise models such as GIPCOL; the performance gap here is larger than in the closed-world case, underscoring the value of explicit primitive integration. \textbf{Visual disentanglement} stands out as the most effective category, with augmentation-based (Retrieval), prototype-centric (CLUSPRO), conditional (CDS-CZSL), and auxiliary (VisPrompt) approaches all achieving strong results. Notably, \textit{Retrieval dominates on CGQA}, suggesting that incorporating training knowledge is especially beneficial on larger datasets, where the test space expands significantly and models must navigate a vast number of compositions. \textit{On MIT-States, CLUSPRO emerges as the top performer, while on UT-Zappos, it again secures the best results}. By contrast, \textbf{cross-modal approaches}, which were competitive with top visual methods in the closed-world setting, fall behind in open-world evaluation; even the strongest model, CAILA, trails top visual methods such as Retrieval and CLUSPRO. Overall, these results from Table \ref{tab:results} and Figures (\ref{fig:scatter-mit-open}, \ref{fig:scatter-ut-zappos-open}, \ref{fig:scatter-cgqa-open}) highlight that \textit{while cross-modal methods show strong potential, visual disentanglement currently provides more reliable performance under the expanded open-world label space}, and further research is needed to fully exploit cross-modal strategies.

\section{Open Challenges \& Future Directions}
\label{open-challenges-future-direction}
While our taxonomy and discussion on results highlight the significant progress achieved across no disentanglement, textual, visual, and cross-modal disentanglement approaches, they also reveal persistent gaps that shape the frontier of CZSL research. Each category tackles aspects of discriminability and contextuality in different ways. Their limitations, whether in handling attribute variability, managing unfeasible compositions, generalizing to unseen primitives, or scaling to large multimodal models, point directly to the open challenges ahead. Building on these observations, we now outline the most pressing research directions, implicitly linked to the strengths and weaknesses identified in our taxonomy and comparative analysis.

\subsection{Modeling primitives and contextuality}
Accurately capturing the dynamic nature of attributes remains a central challenge across all categories of CZSL. Early no-disentanglement methods fail to capture contextuality altogether, while textual disentanglement, despite leveraging linguistic separability, struggles to provide a true understanding of primitive variability. Visual disentanglement has delivered the strongest gains so far, with several methods achieving state-of-the-art results, yet even here explicit separation can oversimplify attribute-object interactions. Cross-modal approaches are especially promising, as they combine linguistic structure with visual grounding and already surpass many baselines, yet comparative evaluations show they still do not consistently match the strongest visual disentanglement methods, indicating that their full potential for robust compositional generalization has not yet been realized. Future research should therefore continue to refine visual disentanglement strategies while placing greater emphasis on developing cross-modal frameworks, which hold the greatest potential for capturing contextuality in a scalable and robust way.

\subsection{Scaling to open-world evaluation}
Our comparative analysis highlight a sharp performance drop when moving from closed-world to open-world evaluation, where the label space expands to include the full cross-product of attributes and objects, encompassing both feasible and infeasible compositions. Existing approaches attempt to mitigate this issue by leveraging external knowledge sources or correlations observed in the training data, using them to either prune unfeasible pairs or downweight them through auxiliary losses. While such mechanisms provide incremental improvements, they remain limited: they require additional resources and, more importantly, sidestep the core problem rather than solving it. In principle, a truly generalizable CZSL model should not be confused by unfeasible compositions. For example, a system capable of recognizing ``red apple'' should naturally assign negligible probability (or score) to an infeasible composition like ``hairy apple.'' Achieving this would demonstrate genuine compositional reasoning rather than reliance on feasibility filters. Therefore, a key direction for future work is the development of models whose representations are intrinsically robust to unfeasible compositions, narrowing the gap between closed-world and open-world performance without the need for explicit feasibility calculations.

\subsection{Generalization to unseen primitives}
Although standard CZSL assumes a fixed vocabulary of attributes and objects, real-world scenarios inevitably involve encountering entirely new primitives. Recent works have introduced new settings such as open-vocabulary CZSL \cite{OV-CZSL}, partial supervision \cite{kg-sp}, and incremental primitive learning \cite{zhang2024continualCZSL} to explore these more challenging scenarios, but only a handful of methods have been proposed within them, and their performance remains well below standard closed-world and open-world benchmarks. To advance the field, future work should not only design models for these scenarios but also routinely evaluate them alongside the conventional protocols, ensuring robustness across all settings. Promising directions include leveraging visual disentanglement to dynamically adapt attribute representations to unseen objects and exploring cross-modal disentanglement to exploit the semantic expansion capacity of language models. Addressing these settings systematically is essential for developing CZSL systems that can scale to the evolving and open-ended nature of real-world data.

\subsection{Leveraging large multimodal models}
Generalization in CZSL ultimately depends on learning robust primitive representations, making the quality of both visual and textual embeddings crucial. Early approaches that relied on CNN-based visual encoders consistently underperformed, whereas our comparative analysis show that replacing them with CLIP backbone led to substantial improvements in accuracy across taxonomy categories. This clear jump suggests that stronger encoders directly enable better disentanglement and compositional generalization. A natural next step is to explore Large Multimodal Models (LMMs) also known as Multimodal Large Language Models (MLLMs), which combine powerful LLMs with web-scale multimodal pretraining. Architectures such as LLaVA \cite{liu2023llava}, MiniGPT-4 \cite{zhu2023minigpt}, and Qwen-VL \cite{Qwen-VL} exhibit advanced visual reasoning and semantic alignment, offering the potential to generate richer and more context-sensitive attribute-object representations than current vision-language models. Combining cross-modal models with the richer visual-textual embeddings provided by LMMs may offer a promising path toward stronger compositional generalization, enabling models to move beyond alignment alone toward more robust and context-sensitive reasoning.

Despite their promise, however, LMMs also raise critical challenges. A primary concern is data contamination: because they are trained on massive web corpora, it is difficult to guarantee that benchmark datasets such as MIT-States, UT-Zappos, or CGQA were not included in pretraining, risking inflated results that reflect memorization rather than genuine compositional generalization. Furthermore, their size and complexity exacerbate the efficiency challenges already seen with Clip-based adaptation. Future research should therefore pursue two directions: (i) establishing rigorous evaluation protocols that account for contamination, possibly through the creation of clean or synthetic benchmarks, and (ii) developing adaptation strategies that exploit LMMs’ reasoning power while constraining them to learn true compositional structures rather than surface correlations. In doing so, LMMs could provide the foundation for bridging the current gap between strong visual disentanglement and robust cross-modal generalization. \\

 \textbf{Summary:} While significant progress has been made in compositional zero-shot learning, our taxonomy and empirical analysis reveal that achieving truly robust and generalizable recognition across diverse settings remains an open problem. Each family of methods contributes partial solutions, textual disentanglement leverages linguistic separability but struggles with visual variability, visual disentanglement achieves strong performance yet remains brittle for subtle attributes, and cross-modal approaches hold great promise but have yet to fully realize their potential, particularly in open-world evaluation. The challenges outlined, ranging from nuanced modeling of attribute contextuality, handling large and unfeasible label spaces, and scaling to computationally efficient and contamination-free large multimodal models and to adapting dynamically to unseen primitives, underscore both the depth and breadth of this research frontier. Addressing these challenges will require innovation that cuts across categories: developing context-adaptive attribute models, training objectives that encourage robustness without relying on explicit feasibility calculations, and scalable, efficient architectures. Successfully overcoming these obstacles is crucial for unlocking the full potential of CZSL, enabling intelligent systems to reason about compositions with a flexibility and transparency approaching human cognition.

\section{Conclusion}
This survey presented the first comprehensive taxonomy of Compositional Zero-Shot Learning (CZSL), organizing methods into four families: no disentanglement, textual, visual, and cross-modal (hybrid) disentanglement, based on how they address the core challenges of discriminability and contextuality. Our comparative evaluation across closed-world and open-world settings shows that CLIP-based backbones consistently outperform ResNet, visual disentanglement remains the most effective category, and cross-modal approaches, though promising, still trail behind under open-world conditions. Building on these insights, we identified open challenges in modeling attribute contextuality, managing unfeasible compositions, adapting to unseen primitives, and leveraging LMMs efficiently. We hope this survey serves both as a synthesis of past progress and as a roadmap for future research, guiding the development of CZSL models that combine scalability, robustness, and transparency in reasoning about novel compositions.


\bibliographystyle{ACM-Reference-Format}
\bibliography{sample-base}

\appendix





\end{document}